\documentclass[10pt,onecolumn]{IEEEtran}

\PassOptionsToPackage{numbers, compress}{natbib}
\usepackage[utf8]{inputenc} 
\usepackage[T1]{fontenc}    
\usepackage{hyperref}       
\usepackage{url}            
\usepackage{booktabs}       
\usepackage{amsfonts}       
\usepackage{nicefrac}       
\usepackage{microtype}      
\usepackage[noadjust]{cite}
\usepackage{amsmath}
\usepackage{multirow}

\DeclareMathAlphabet\mathbfcal{OMS}{cmsy}{b}{n}
\usepackage[tight,footnotesize]{subfigure}
\usepackage{amsthm}
\usepackage{bbm}
\usepackage{algorithm}
\usepackage{algorithmic}
\usepackage{graphicx,amsmath,amssymb,latexsym,epsfig}
\usepackage{setspace}
\usepackage{psfrag}

\usepackage{lscape}

\newtheorem*{proof of Theorem*}{Proof of Theorem 3}
\newtheorem{proof of Lemma}{Proof of Lemma}

\title{Deep Learning with the Random Neural Network and its Applications}

\author{Yonghua Yin\\
Intelligent Systems and Networks Group,
Electrical \& Electronic Engineering Department\\
Imperial College, London SW7 2AZ, UK\\
 \texttt{y.yin14@imperial.ac.uk, yinyongh@foxmail.com} \\
}

\begin{document}

\maketitle

\begin{abstract}
The random neural network (RNN) is a mathematical model for an ``integrate and fire'' spiking network that closely resembles the stochastic behaviour of neurons in mammalian brains. Since its proposal in 1989, there have been numerous investigations into the RNN's applications and learning algorithms. Deep learning (DL) has achieved great success in machine learning. Recently, the properties of the RNN for DL have been investigated, in order to combine their power. Recent results demonstrate that the gap between RNNs and DL can be bridged and the DL tools based on the RNN are faster and can potentially be used with less energy expenditure than existing methods.
\end{abstract}

\section{Introduction}

Originally developed to mimic the behaviour of the brain's biological neurons, the random neural network (RNN) proposed by E. Gelenbe,  functions as a stochastic integer state ``integrate-and-fire'' system  \cite{CRAS,Stable}, and was detailed in early papers \cite{RNN89,RNN90,RNN93}.
In an RNN, an arbitrarily large set of neurons interact with each other via excitatory and inhibitory spikes that modify each neuron's action potential in continuous time.
It has been proved that, in a steady state, the stochastic spiking behaviours of the RNN have a remarkable property called ``product form'', with the state probability distribution given by an easily-solvable system of nonlinear equations.
Compared to many other neural network models, e.g., the multi-layer perceptron (MLP) \cite{Rumelhart1986},
the radial basis function (RBF) neural network \cite{park1993approximation},
extreme learning machines (ELM) \cite{Cambria2013ELM}
and orthogonal polynomial networks (OPN) \cite{zhang2014cross}, the RNN has the following advantages:\\
1) The RNN model is closer to the biological neuronal network and can better represent how signals are transmitted through the human brain, whereby the complex stochastic behaviours among neurons can be modelled and analysed more easily in a mathematical form.\\
2) The RNN model generally has a better predictive capability because of its non-negativity and probability constraints.\\
3) The RNN model has rigorously established mathematical properties which simplify the related computations.\\
4) Owing to the highly-distributed nature, the RNN model can potentially be deployed in either energy-efficient customised or neuromorphic hardware.

The RNN has been used to model and explain that cortico-thalamic oscillations observed in the rat somatosensory system are caused by positive feedback in cortex  \cite{Biosystems} and is an example of a mathematical model for natural computation \cite{Natural} that has also been used for protein alignment \cite{PhanSG12}.

The RNN has many engineering applications: it represents
 image textures \cite{Atalay} to obtain accurate segmentation of brain Magnetic Resonance Images \cite{MRI}. It achieves high video compression ratios  at low computational cost \cite{Cramer1,Cramer2,Cramer3}, and has been used for vehicle classification \cite{abdelbaki2001laser}. As a fast heuristic for task assignment in distributed systems \cite{Aguilar}, it also optimises a network's connectivity \cite{RNN6} and can also be used to generate error correcting codes \cite{abdelbaki1999random}. With reinforcement learning, it improves the Quality of Service experienced by network users \cite{CPN,Brun,Francois1,Francois2}, by adapting the routing to the traffic characteristics of the traffic in packet networks \cite{Zarina,Wang2016}, to improve web search \cite{Web}, to dynamically allocate tasks in Cloud Computing systems \cite{Wang2}, to predict the toxicity of compounds from data on their physical-chemical characteristics \cite{ICANN18}, and to detect network attacks \cite{sakellari2010demonstrating,IoT}.

The RNN is actually a special case of a class of G-Networks, a general purpose probability models for large scale networks \cite{GN1,GN2,GN3,GN4,GN5,GN6} which were shown to have the desirable product form solution \cite{Introduction}  to efficiently compute their mathematical characteristics. While all the potential of such models for machine learning has still not been exploited \cite{gelenbe1994g}, G-Networks have had applications in areas different from neuronal networks, such as for load balancing in distributed systems and communication networks \cite{Morfo}, to model Gene Regulatory Networks \cite{GelenbeRegNets2007,KimG12}, and to represent systems that operate with intermittent sources of energy \cite{Ceran2,Marin,NOLTA,Kadi}.

Inspired by the structure and function of the brain, deep learning functions as a machine-learning approach \cite{hinton2006reducing,lecun2015deep}.
Most computational models in deep learning exploit a feed-forward neural-network architecture composed of multi-processing layers; this architecture allows the model to extract high-level representations from the raw data. In this manner, essential and useful information can be extracted from this raw data so that the problem of building complex mapping from the raw inputs to the desired outputs becomes more solvable. Deep learning has achieved major advances in solving many historic issues in machine learning.

Since the proposal of the RNN in 1989, there have been numerous investigations into the RNN's applications and learning algorithms \cite{rnn2000,rnn2010,GEORGIOPOULOS2011361}; however, there has not been any research regarding the property of the RNN for deep learning until the first related work of the authors in 2016 \cite{gelenbedeep2016}. The main objective of the following work of the authors in \cite{gelenbedeep2016,gelenbedeep2016_SAI,yin2016deep,
yin_cloud2017,2016arXiv160908151Y,gelenbe2017deep,yyh_rnnl_ijcnn2017} is to connect the RNN with deep learning, which is novel in the following two aspects: 1) the research could provide powerful and deep-learning tools that not only achieve state-of-the-art performance but can potentially be used with less energy expenditure than existing methods;
2) the research attempts to provide a better understanding of the relationship between deep learning and the brain.
The details of the progresses in this direction has been summarized into the PhD thesis of the author Yonghua Yin \cite{phd-yonghua2018}.

This paper is organized into the following sections.\\
1) Section \ref{chap.literature} presents the backgrounds of neural networks, deep learning and  the RNN.\\
2) Section \ref{chapter.rnn_classifier} illustrates the classifier based on the RNN function approximator \cite{gelenbe1999function,99RNN_appro,gelenbe2004function}.\\
3) Section \ref{chapter.lrnn} illustrates the multi-layer non-negative RNN autoencoders proposed in \cite{2016arXiv160908151Y} for dimension reduction, which are robust and effective in handling various types of data.\\
4) The work on dense random neural network \cite{gelenbedeep2016,gelenbedeep2016_SAI,yin2016deep,gelenbe2017deep} that includes mathematical modelling and deep-learning algorithms have been summarised in Section \ref{chapter.densernn}.\\
5) Section \ref{chapter.mlrnn} presents the investigation into the standard RNN and demonstrates its power for deep learning \cite{yyh_rnnl_ijcnn2017}. The resultant deep-learning tool is demonstrated to be effective and is arguably the most efficient of the five different deep-learning approaches.\\
6) The conclusions and future work are given in Section \ref{ch:conclusions}.

\section{Background} \label{chap.literature}

\subsection{Neural Networks} \label{sec.ANN}

In 1943, McCulloch and Pitts \cite{McCulloch1943} built mathematical models of neural networks to describe how the neurons in the brain might work, where the neural behaviours are treated by means of propositional logic.
This is the first computational model of neuron activity.
In \cite{Rochester1956}, Rochester conducted two sets of simulation experiments representing neural networks with 69 and 512 neurons, respectively, using digital computers, in order to test theories on how the brain works.
The theory in \cite{rosenblatt1958perceptron} was developed for a hypothetical nervous system, called a perceptron, which detailed the organisation of a perceptron in the terms of impulse types, connection patterns and so on, and the mathematical analysis of learning in the perceptron.

In 1989, Hornik \cite{hornik1989multilayer} proved that a multi-layer feed-forward neural network, in other terms, multi-layer perceptron (MLP), with as few as one hidden layer can approximate any continuous function uniformly, provided that the hidden-layer neurons use non-decreasing squashing functions $\mathbb{R} \rightarrow [0,1]$, e.g., sigmoid.
The hidden-layer output is of the form
$\sigma(\sum_{i=1}^{N}(\beta_i x_{i}+b_{i}))$, where $x_i$, $N$, $\beta_i$, $b_i$ and $\sigma(\cdot)$ are respectively the input, input dimension, connection weight, threshold and activation function. This form is called the ridge function (RF) \cite{scarselli1998universal}.
Stinchcombe and White \cite{Stinchcombe1989} showed that sigmoid functions are not necessary for universal approximation.
Further work by \cite{hornik1991approximation} investigated the approximation capability of the MLP with bounded and nonconstant activation functions.
A generalised result from \cite{hornik1991approximation} in \cite{leshno1993multilayer} relaxes the condition, i.e., an MLP with non-polynomial activation functions is a universal approximator.

In addition to the RF-based neural networks, there are other types. The radial basis function (RBF) based neural networks are also widely used, which have the hidden-output form $\sigma(||X-C||/a)$, where $X \in \mathbb{R}^{N \times 1}$ is the input, $C \in \mathbb{R}^{N \times 1}$ and $a$ are respectively the centre and impact factor of the hidden node. \cite{park1991universal,park1993approximation}  proved the universal approximation property of the RBF-based neural networks.
Another type of neural network uses a product-based orthogonal-polynomial (POP) activation function in \cite{zhang2014cross,zhang2014weights}. The hidden-output form is $\prod_{i=1}^{N} p_{i}(x_i)$, where $p_{i}(\cdot)$ is an orthogonal polynomial. The POP neural networks are also universal approximators. It is also proved that the POP-based single-hidden-layer neural network (SLNN) in \cite{zhang2014cross} has lower computational complexity than the RF-based one.

With recent improvements, back propagation (BP)
has become a standard method to train neural networks.
For example, the ReLU units make neural networks easier to train using BP \cite{AISTATS2011_GlorotBB11}.
Weight-sharing strategies, e.g., the type used in convolutional neural networks, reduce the number of parameters needed to be trained, which facilitates the training of multi-layer neural networks.
Training a neural network with BP well can be time-consuming and require huge computation load because of the possible slow convergence and potential instability inherent in the training process.
This issue has provided the rationale for recent research focusing on the subject of neural networks that adopted the perspectives of both the approximation property and training efficiency.

Huang \cite{huang2006extreme} investigated RF-based neural networks and pointed out that most results on their universal approximation properties are based on the assumption that all weights and biases (including $\beta_i$ and $b_i$) need to be adjusted.
As a result, the training of these networks must proceed at a slower pace.
Huang \cite{huang2006extreme} proved that, using randomly-generated $\beta_i$ and $b_i$, a RF-based single-hidden-layer feed-forward neural network (SLFN) with $N$ hidden nodes can learn $N$ distinct samples exactly. Therefore, the SLFN can be described as a linear system whose output weights can be analytically determined. This is the extreme learning machine (ELM) concept. Learning systems based on the ELM can be trained more than a hundred times faster than the BP-based ones without compromising accuracy in many cases \cite{mlelm}.
Zhang \cite{zhang2014cross,zhang2014weights} focused on the POP neural network. Theoretical results in \cite{zhang2014cross,zhang2014weights} present its universal approximation property. Similar to the ELM concept, the POP neural network can then be described as a linear system with the output weights analytically determined, which is the weights-direct-determination (WDD) concept.

\subsection{Spiking Neural Network and Neuromorphic Computing} \label{sec.snn_computing}

Von Neumann architecture \cite{Burks_von_Neumann2008} has provided the basis for many major advances in modern computing \cite{cheng2016exploring,wen2016new}, allowing for the rapid decrease in transistor size, and improved performance without increasing costs. However, the rate of progress experienced by this type of architecture has recently slowed down, and with various seemingly insurmountable obstacles preventing further significant gains, the end of Moore's Law is in sight \cite{pavlus2015search}. One major obstacle is that, due to quantum-mechanical effects, transistors smaller than 65 nanometres can no longer switch between ``on'' and ``off'' reliably, which means that a digital computer cannot distinguish one from zero. The frequency wall is also a major obstacle, which means that the hardware could melt down because of excess heat if the CPU frequency exceeds four billion times per second. Therefore, a new machine that goes beyond the boundaries imposed by von Neumann architecture needs to be developed, and the areas of neuromorphic computing and spiking neural network (SNN) are starting to attract significant attention from researchers in both academic and commercial institutions.

A non-von Neumann architecture is the TrueNorth cognitive computing system, which has been developed by IBM and is inspired by how organic brains handle complex tasks easily while being extremely energy-efficient \cite{esser2013cognitive,preissl2012compass,dominguez2016multilayer}. The TrueNorth chip has received a lot of attention from wider society since it is the first large-scale commercial spiking-based neuromorphic computing platform \cite{cheng2016exploring}.
A first major step occurred when IBM researchers presented an architectural simulator, called Compass, for large-scale network models of TrueNorth cores \cite{preissl2012compass}. Since the Compass simulator is functionally equivalent to TrueNorth and runs on a von Neumann computer, it facilitates research and development for TrueNorth at this stage.
Based on this simulator, in \cite{esser2013cognitive}, applications - including speaker recognition, digital recognition and so on - are integrated to demonstrate the feasibility of ten different algorithms, such as convolution networks, liquid state machines and so on, to be run in TrueNorth.
The conventional neural network algorithms require high precision while the synaptic weights and spikes in TrueNorth have low quantisation resolution, which is the issue investigated in  \cite{wen2016new,cheng2016exploring}. Considering that TrueNorth is spiking-based, naturally, mapping models of spiking neural networks into TrueNorth could be a reasonable choice \cite{wen2016new,cheng2016exploring}. In \cite{wen2016new} and \cite{cheng2016exploring}, a single-hidden-layer neural network was deployed into TrueNorth with the learning algorithm presented, where the nonlinearity is brought by the cumulative probability function.
The research in \cite{diehl2016truehappiness} implemented a sentiment-analysis algorithm on both the spiking-neural-network simulator and TrueNorth system. First, a fully-connected single-hidden-layer neural network (FCNN) with rectified linear units was trained using the stochastic gradient descent. Then, this FCNN was converted into a SNN model with a discussion on the related issues needed to be addressed, e.g., the precision and the negative inputs. There is a performance drop in this conversion. Finally, the SNN was mapped into TrueNorth. It was mentioned that performance loss in this step is more significant than that through the FCNN-SNN conversion.

The spiking neural network architecture (SpiNNaker) project is another spike-based neuromorphic computing platform \cite{furber2014spinnaker,knight2016efficient,dominguez2016multilayer}. The SpiNNaker aims to model a large-scale spiking neural network in biological real time. The paper \cite{furber2014spinnaker}, published in 2014, presented an overview of the SpiNNaker architecture and hardware implementation, introduced its system software and available API and described typical applications.
A research project on SpiNNaker in \cite{knight2016efficient} presented an application for audio classification by mapping a trained one-hidden-layer leaky integrate-and-fire SNN into the SpiNNaker, utilising the PyNN library \cite{davison2009pynn} that is a common interface for different SNN simulators including NEURON \cite{carnevale2006neuron}, NEST \cite{Gewaltig:NEST} and Brian \cite{goodman2008brian}.

Recently, SNN-based neuromorphic computing has made rapid progress by combining its own inherent benefits and deep learning \cite{esser2016convolutional}. By enforcing connectivity constraints, the work in \cite{esser2016convolutional} applied the deep convolutional network structure into neuromorphic systems. Conventional convolutional networks require high precision while neuromorphic design uses only low-precision one-bit spikes. However, it was demonstrated that, by introducing constraints and approximation for the neurons and synapses, it is possible to adapt the widely-used backpropagation training technique. This enables the creation of a SNN that is directly implementable into a neuromorphic system with low-precision synapses and neurons \cite{esser2015backpropagation}. Comparisons based on 8 typical datasets demonstrated that this neuromorphic system achieves comparable or higher accuracies compared with start-of-the-art deep learning methods. High energy efficiency is maintained throughout this process.

\subsection{Deep learning}

Deep learning \cite{hinton2006reducing,lecun2015deep} has achieved great success in machine learning and is currently contributing to huge advances in solving complex practical problems compared with traditional methods. For example, based on deep convolutional neural networks (CNN), which are powerful in handling computer vision tasks, deep learning systems could be designed to realise the dream of self-driving cars \cite{2016arXiv160407316B}. Another example is the game of Go, invented in ancient China more than 2,500 years ago, which was one of the most challenging classic games for artificial intelligence since the number of possible moves is greater than the total number of atoms in the visible universe \cite{Go}. However, a major step was recently taken when a computer system based on deep neural networks, a product of the Google DeepMind laboratories, defeated some of the world's top human players \cite{silver2016mastering}.

Many computational models used in deep learning exploit a feed-forward neural-network architecture that is composed of multi-processing layers, which allows the model to extract high-level representations from raw data. The feed-forward fully-connected multi-layer neural network, also known as multi-layer perceptron (MLP), has already been investigated for decades \cite{rosenblatt1961principles}. However, progress on the MLP was slow since researchers found difficulties in training MLPs with more than three layers \cite{glorot2010understanding}. Furthermore, adding more layers did not seem to help improve the performance. Pre-training the MLP layer by layer is a great advance \cite{hinton2006reducing,hinton2006fast}, and this breakthrough demonstrates that it is possible to train the MLP and adding more layers can actually improve the network performance. This method is still very useful due to its wide adaptability. Recent literature shows that utilising the rectified linear unit (ReLU) reduces the difficulty in the training of deep neural networks \cite{AISTATS2011_GlorotBB11}. The typical training procedure called stochastic gradient descent (SGD) provides a practical choice for handling large datasets \cite{bousquet2008tradeoffs}.

A considerable body of successful work has been amassed thanks to LeCun's inspirational work on CNN \cite{mnist}.
The weight-sharing property in the convolutional structure significantly reduces the number of weights to be learned and the local connectivity pattern allows good feature extraction, which facilitates the training of a deep CNN. Recent work on deep residual learning has made it possible to train extremely deep neural networks with thousands of layers \cite{he2016deep}.

The task of supervised learning in the DNN can be handled using the greedy layer-wise unsupervised learning, convolution structures and residual learning. Given the ease with which humans are able to learn in an unsupervised manner, and the ready availability of raw datasets - which is not the case with labelled datasets - unsupervised learning has become a key issue upon which the deep learning community has begun to focus. A popular topic in unsupervised learning is the generative adversarial nets (GAN) system \cite{NIPS2014_5423}, which includes a discriminative and generative models. The generative model that takes noise as input tries to generate fake samples to cheat the discriminative model such that it cannot distinguish between fake and true samples. The GAN can be used to model natural images and handle image generation tasks \cite{DBLP:journals/corr/RadfordMC15}. The adversarial autoencoders (AAE) are proposed based on the GAN \cite{DBLP:journals/corr/MakhzaniSJG15}. The work in \cite{DBLP:journals/corr/MakhzaniSJG15} shows how the AEE can be used in different applications, e.g., semi-supervised classification, unsupervised clustering, dimensionality reduction and data visualisation.

\subsection{Random neural network}

The random neural network (RNN) proposed by E. Gelenbe in 1989 \cite{RNN89,RNN90,RNN93} is a
stochastic
integer-state ``integrate and fire'' system
and was developed to mimic the behaviour of biological neurons in the brain. The RNN is also a recurrent neural network.
In an RNN, an arbitrarily large set of neurons interact with each other via excitatory
and inhibitory spikes which modify each neuron's action potential in
continuous time. The power of the RNN is derived from the fact that, in steady state, the stochastic spiking behaviors of the network have a remarkable property
called ``product form'' and that the state probability distribution is given by an easily solvable system of non-linear equations.
In addition, the RNN models the biological neuronal network with greater accuracy, and offers an improved representation of signal transmission in the human brain, when compared to alternative neural network models, e.g., the MLP \cite{rumelhart1985learning}, ELM \cite{huang2006extreme} and OPN \cite{zhang2014cross,zhang2014weights}.

\subsubsection{Model Description} \label{sec.chp_intro_rnn}

Within the RNN system with $L$ neurons, all neurons communicate with each other with stochastic unit amplitude spikes while receiving external spikes. The potential of the $l$-th neuron, denoted by $k(t) \geq 0$, is dynamically changing in continuous time, and the $l$-th neuron is said to be excited if its potential $k(t)$ is larger than zero. An excitatory spike arrival to the $l$-th neuron increases its potential by 1, denoted by $k(t^+) \leftarrow k(t)+ 1$; while an inhibitory spike arrival decreases its potential by 1 if it is larger than zero, denoted by $k(t^+) \leftarrow \max(k(t) - 1, 0)$, where $\max(a,b)$ produces the larger element between $a$ and $b$.

\begin{figure}[t]
\centering
\includegraphics[width=2.2in]{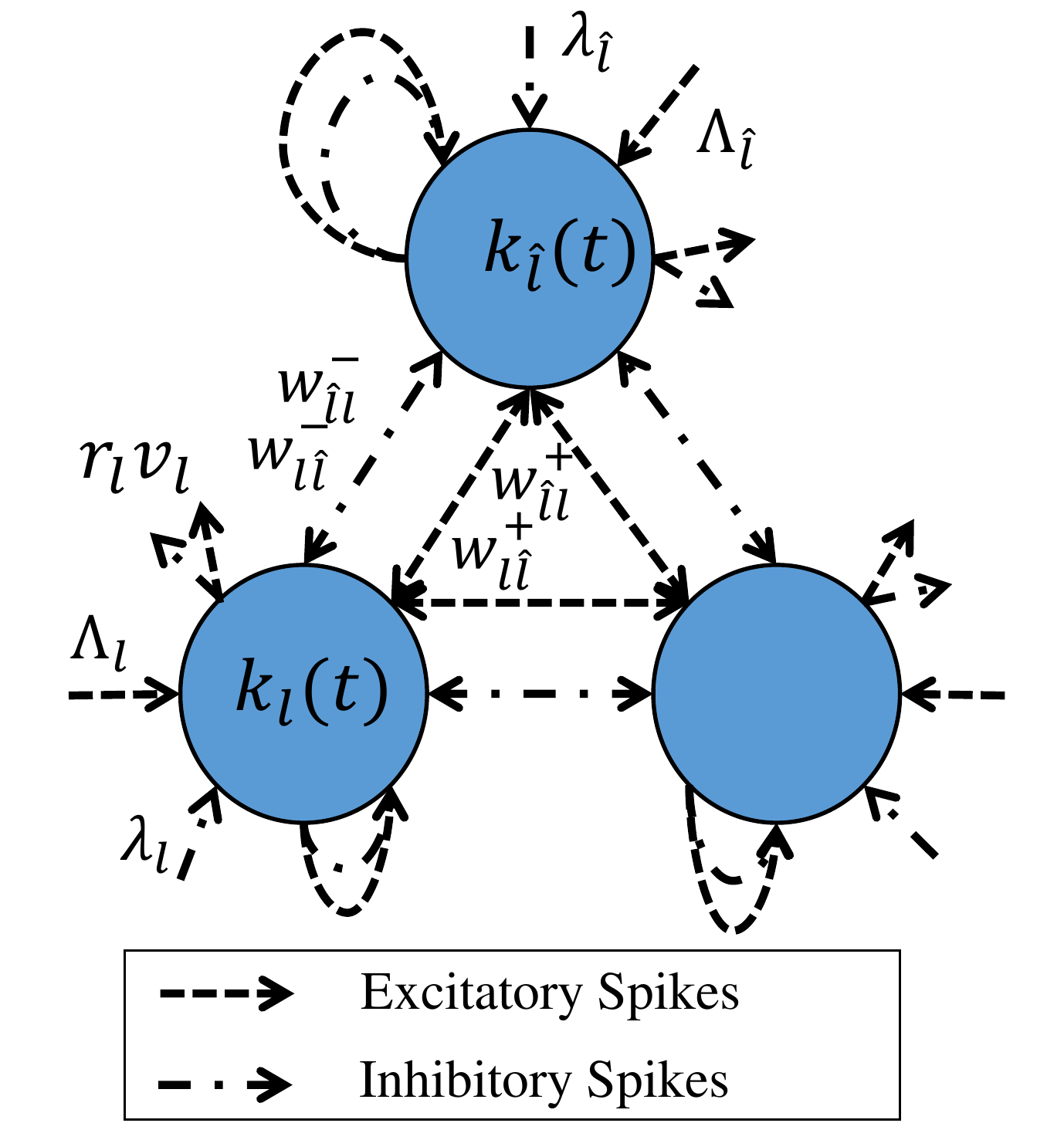}
\caption{Schematic representation of a spiking RNN system.}
\label{fig.chp2RNNModel}
\end{figure}

For better illustration, Figure \ref{fig.chp2RNNModel} presents the schematic representation of the RNN system.
Excitatory spikes arrive at the $l$-th neuron from the outside world according to Poisson processes of rate $\Lambda_l$, which means that the probability that there are $\varrho$ excitatory spike arrivals in time interval $\Delta t$ is \cite{heeger2000poisson}:
$
\text{Prob}(\varrho~\text{spike~arrivals~in~interval}~\Delta t) = {(\Lambda_l \Delta t)^{\varrho} e^{-\Lambda_l \Delta t}}/{\varrho!},
$
where $\varrho$ is a non-negative integer.
In addition, the rate of inhibitory Poisson spike arrivals from the outside world to the $l$-th neuron is $\lambda_{l}$.
When the $l$-th neuron is excited, it may fire excitatory or inhibitory spikes towards other neurons with the inter-firing interval $\bar{\varrho}$ being exponentially distributed, which means that the probability density function of $\bar{\varrho}$ is
$
\text{Prob}(\bar{\varrho}=\Delta t) =  r_l e^{- r_l \Delta t},
$
where $r_l$ is the firing rate of the $l$-th neuron. When the $l$-th neuron fires a spike, its potential is decreased by 1.
The fired spike heads for the $\hat{l}$-th neuron as an excitatory spike
with probability $p^{+}_{l,\hat{l}}$ or as an inhibitory spike with
probability $p^{-}_{l,\hat{l}}$, or it departs from the network/system with
probability $\nu_l$. The summation of these probabilities is 1: $\sum_{\hat{l}=1}^{L}(p^{+}_{l,\hat{l}}+p^{-}_{l,\hat{l}})+\nu_l=1.$

Evidently, the potentials of the $L$ neurons in the system are dynamically changing over time due to the stochastic spikes and firing events.
Let $\text{Prob}(k_{l}(t)>0)$ denote the probability that the $l$-th neuron is excited at time $t$. Accordingly, let $q_l=\lim_{t\rightarrow \infty} \text{Prob}(k_{l}(t)>0)$ denote stationary excitation probability of the $l$-th neuron. Due to the stochastic and distributed nature of the behaviours of the whole spiking neural system, it is difficult to obtain the value of $q_l$. For a system with fixed configurations and inputs, a straightforward method is to estimate the value of $q_l$ by using the Monte Carlo method. However, this method may not enable us to obtain a good estimation of $q_l$ or be applicable when the number of neurons becomes very large.

In \cite{RNN89}, Gelenbe presented important results on the excitation probabilities of the neurons of this RNN system. It is proven in \cite{RNN89,RNN90,RNN93} that $q_l=\lim_{t\rightarrow \infty} \text{Prob}(k_{l}(t)>0)$ can be directly calculated by the following system of equations:
\begin{equation}\label{eqn.rnn}
q_l=\min(\frac{\lambda^{+}_l}{r_l + \lambda^{-}_l},1),
\end{equation}
where $\lambda^{+}_l=\Lambda_l+\sum_{\hat{l}=1}^{N}{q_{\hat{l}} w^{+}_{\hat{l},l}}$, $\lambda^{-}_l=\lambda_l+\sum_{\hat{l}=1}^{N}{q_{\hat{l}} w^{-}_{\hat{l},l}}$, $w^{+}_{\hat{l},l}=r_{\hat{l}} p^{+}_{\hat{l},l}$, $w^{-}_{\hat{l},l}=r_{\hat{l}} p^{-}_{\hat{l},l}$ and $l=1,\cdots,L$. Here $\Lambda_l$ and $\lambda_{l}$ are respectively the arrival rates of external excitatory and inhibitory spikes and $r_l$ is the firing rate of the $l$-th neuron.
In addition, $\sum_{\hat{l}=1}^{L}(p^{+}_{l,\hat{l}}+p^{-}_{l,\hat{l}})+\nu_l=1$.
Operation $\min(a,b)$ produces the smaller one between $a$ and $b$. In \cite{RNN93}, it was shown that the system of $N$ non-linear equations (\ref{eqn.rnn}) have a solution which is unique.
Therefore, the states of the RNN can be efficiently obtained by solving a system of equations without requiring the Monte Carlo method \cite{monte_carlo}.

\subsubsection{Learning in Random Neural Network}

Solving supervised learning problems is a key issue in machine learning and artificial intelligence \cite{nnlearning,nnlearning2}, as this is a requirement for the majority of practical applications of neural networks. In general, supervised learning of a system requires it to learn mapping from the input data (integers or real values) to the output data (labels or real values) in a given training dataset. Investigations have been carried out to facilitate the development of the RNN learning methods since 1993 \cite{RNN93}, with the goal of solving the RNN's learning problems. In the work of \cite{RNN93}, Gelenbe developed a gradient-descent learning algorithm for the RNN, which seeks non-negative weights in terms of the gradient descent of a quadratic cost function and is the most widely-used learning algorithm for the RNN. Likas \cite{RNN_BP_improve} utilises quasi-Newton methods and the derivative results of gradient descent in \cite{RNN93} for the RNN learning. The proposed algorithm was tested on the RNN with two different architectures, where the fully recurrent architecture yielded poor results while the feed-forward architecture yielded smaller errors with fewer steps than the pure gradient-descent algorithm. It is also reported that the proposed algorithm has higher computational complexity and cannot be applied to online learning.

Based on \cite{RNN93}, Basterrech \cite{RNN_BP_improve2} exploited the Levenberg-Marquardt optimisation procedure and adaptive momentum. Its performance was evaluated when applied to several problems. Compared to the gradient-descent algorithm, it achieved faster convergence speed in almost all cases but proved less robust in solving the XOR problem. The RPROP learning algorithm introduced in \cite{RPROP}, which performs a direct adaptation of the weight step based on the local gradient information, is combined with the gradient-descent algorithm \cite{RNN93} for the RNN learning \cite{RNN_BP_improve2}. Georgiopoulos gave a brief description and some critical comments on this RPROP algorithm in \cite{learn_rnn}. It is reported in \cite{RNN_BP_improve2} that, in the learning of the geometrical images, this RPROP algorithm outperformed the gradient-descent algorithm. The gradient-descent-based learning algorithms may have some inherent weaknesses, such as slow convergence and being trapped into poor local minima. Instead of using the gradient descent, Georgiopoulos applied the adaptive inertia weight particle swarm optimization (AIW-PSO) \cite{rnn_pso} and differential evolution (DE) approach \cite{rnn_de} for the RNN learning in \cite{learn_rnn}.
Then, the performances of the gradient-descent algorithm, RPROP algorithm, AIW-PSO and DE approach for classifying 12 datasets were compared, and there is no clear overall winner among them. In addition, Timotheou tackled the learning problem of the RNN from a different perspective  \cite{rnn2009NNLS,rnn2008NNLS}. Specifically, the learning problem was first approximated to obtain a non-negative least-square (NNLS) problem. The next step is to apply the solution to the NNLS problem directly to the RNN learning problem. It is reported in \cite{rnn2009NNLS,rnn2008NNLS} that this algorithm achieved better performance than the gradient-descent algorithm in a combinatorial optimisation problem emerging in disaster management.

\subsubsection{Applications of Random Neural Network}

The RNN has been successfully used in numerous applications \cite{rnn2000,rnn2010,GEORGIOPOULOS2011361,RADHAKRISHNAN2011347}. In \cite{image_compression}, the RNN is applied in encoding an image into fewer bits in order to achieve image compression by exploiting a feed-forward architecture to learn the mapping from the image to the same image through a narrow passage with fewer hidden-layer neurons, where the extension to video compression applied similar principles \cite{video_compression,video_compression2}. In \cite{recognition_rnn}, multi RNNs are exploited to learn from training data and recognise shaped objects in clutter by combining the results of all RNNs. In \cite{recognition_rnn2}, the feed-forward RNN is applied to learning a mapping from the scanning electron microscopy intensity waveform to the cross-section shape (i.e., the profile) such that the profile can be reconstructed from the intensity waveform and the destruction caused by acquiring a cross-section image can be avoided. In \cite{recognition_rnn3}, the task of Denial of Service (DoS) detection is formulated as a pattern classification problem, and then, with useful input features measured and selected, the RNNs with both the feed-forward and recurrent architectures are exploited to fulfil the task.

Gelenbe proposes the multi-class RNN model in \cite{rnn_multi} as a generalisation following the work of the standard RNN \cite{RNN89}. This multiple-class RNN can be described as a mathematical framework to represent networks that simultaneously process information of various types. Then, by setting the excitatory input rates and the desired outputs of the neuron states as the normalised pixel values, this multiple-class RNN is utilised to learn a given colour texture and generate a synthetic colour texture that imitates the original one \cite{Gelenb02multiclass}. The earlier work in \cite{texture_model} that uses the standard RNN for black and white texture generation utilises similar principles. The RNN learning is also applied to image segmentation \cite{image_segment}, vehicle classification \cite{vehicle_classification} and texture classification and retrieval \cite{texture_classification}. More applications related to utilising the learning capability of the RNN can be seen from the critical review in \cite{learn_rnn}. The many other areas in which the RNN has already been applied are detailed in the reviews in \cite{rnn2000,rnn2010}.

\subsubsection{Implementation of Random Neural Network}

There are three main approaches to implementing the RNN \cite{hardware_rnn}. The most accessible approach is to use computer software tools based on CPU to simulate the RNN spiking behaviours or conduct the RNN equations computations. However, this approach can be of low efficiency because it does not utilise the fact that the RNN can be computed or simulated in a parallel manner. The second approach is still to implement the RNN in computers but in a parallel manner. Specifically, the computations related to the RNN can be conducted simultaneously by a very large number of simple processing elements, where the rapid development of chip multiprocessor (CMP) and Graphics Processing Unit (GPU) make it more feasible. The third approach is to implement the RNN model in hardware using special-purpose digital or analogue components \cite{hardware_rnn}. The RNN's high parallelism can be exploited fully via specially designed hardware, and significant speed increases can be achieved when applying the RNN.

For the first approach, Abdelbaki's work \cite{matlab_rnn} implemented the RNN model using MATLAB software. To the best of the authors' knowledge, no investigation has hitherto been undertaken to implement the RNN using the second approach. In terms of hardware, another of Abdelbaki's studies in \cite{hardware_rnn} shows that the RNN can be implemented by a simple continuous analogue circuit that contains only the addition and multiplication operations. In addition, the Cerkez's study \cite{hardware_rnn2} proposed a digital realisation for the RNN using discrete logic integrated circuits. Kocak's work \cite{hardware_rnn3} implemented the RNN-based routing engine of the cognitive packet networks (CPN) in hardware. This innovation significantly decreased the number of weight terms (from $2n^2$ to $2n$) needing to be stored for each RNN in the engine, where the CPN is an alternative to the IP-based network architectures proposed by Gelenbe \cite{CPN}.

\section{A Classifier based on Random Neural Network Function Approximator}\label{chapter.rnn_classifier}

Neural networks are capable of solving complex learning problems mainly owing to their remarkable approximation capabilities
\cite{lecun2015deep,mlelm}. In the case of the random neural network (i.e., the RNN), the work by Gelenbe in \cite{gelenbe1999function,99RNN_appro,gelenbe2004function}
focused on investigating the approximation property of the RNN. It is proved that the RNN can be constructed as a function approximator with the universal approximation property (UAP).

The authors' work of \cite{RNNClassifier2018} pursues the work of the RNN function approximator in \cite{gelenbe1999function,99RNN_appro,gelenbe2004function} and explores its theoretical framework.
The theorem from \cite{gelenbe1999function,99RNN_appro,gelenbe2004function} is given to show that the RNN has the UAP as a function approximator: for any continuous real-valued function $f(X):
[0,1]^{1 \times N} \rightarrow \mathbb{R}$, there exists an RNN that approximates $f(X)$ uniformly on $[0,1]^{1 \times N}$. The work in \cite{RNNClassifier2018} presents a constructive proof for this UAP theorem.
This lays a theoretical basis for the learning capability
from data of the RNN and the investigation into the capability of the RNN for deep learning in the following sections.
The RNN function approximator is demonstrated to have a lower computational complexity than the orthogonal-polynomial function approximator in \cite{zhang2014cross} and the one-hidden-layer MLP, under certain assumptions.
An efficient training procedure is proposed for the approximator such that it learns a given dataset efficiently, which formulates the learning problem into a convex optimization problem that can be solved much faster than the gradient-descent neural networks.
The RNN function approximator, equipped with the proposed configuration/learning procedure, is then applied as a tool for solving pattern-classification problems. Numerical experiments on various datasets demonstrate that the RNN classifier is the most efficient among the six different types of classifiers, which are the RNN, the Chebyshev-polynomial neural network (CPNN) \cite{zhang2014cross}, ELM \cite{huang2006extreme}, MLP equipped with the Levenberg-Marquardt (LM) algorithm, radial-basis-function neural networks (RBFNN) \cite{wilson1996heterogeneous,yu2009rbf} and support vector machine (SVM) \cite{chang2011libsvm}. The related results can be found in \cite{RNNClassifier2018} and Chapter 3 of \cite{phd-yonghua2018}.

\section{Non-negative Autoencoders with Simplified Random Neural Network}\label{chapter.lrnn}

The authors' work in \cite{2016arXiv160908151Y} combines the knowledge from three machine learning fields, which are: the random neural network (i.e., the RNN), deep learning, and non-negative matrix factorisation (NMF), to develop new multi-layer non-negative autoencoders that have the potential to be implemented in a highly-distributed manner.

\subsection{Technical Background}

The mathematical tool of the RNN has existed since 1989 \cite{RNN89,RNN90,RNN93}, but is
less well known in the machine-learning community. The RNN is developed to mimic the behaviour of biological neurons in the brain.
In the RNN, an arbitrarily large set of neurons interact with each other via stochastic spikes which modify each neuron's action potential in
continuous time. Though the stochastic spiking behaviors are complex, the state probability distribution of the RNN can be calculated by an easily solvable system of non-linear equations. This powerful property of the RNN provides the feasibility of utilizing the techniques from the other machine learning fields, e.g., the deep learning and NMF described as the following, for the development of new learning tools based on the RNN.

Deep learning has achieved great success in machine learning \cite{hinton2006reducing,lecun2015deep}. It is attractive to investigate whether the network architecture and the training techniques in this area could be combined with the RNN or not.
For example. in deep learning, the feed-forward neural-network architecture, composed of multi-processing layers, allows a model to extract high-level representations from raw data. Pre-training a multi-layer network layer by layer is an effective and widely-adaptable technique to tackle the training difficulty in the network \cite{hinton2006reducing,hinton2006fast}. In addition, the typical training procedure called stochastic gradient descent (SGD) provides a practical choice for handling large datasets \cite{bousquet2008tradeoffs}.

Non-negative matrix factorisation (NMF) is also a popular topic in machine learning \cite{lee1999learning,NNPCA,hoyer2002non,wang2013nonnegative,ding2006orthogonal}, and it provides another perspective of optimizing the parameters of the RNN with the constraints of non-negativity.
Lee \cite{lee1999learning} suggested that the perception of the whole in the brain may be based on the part-based representations (based on the physiological evidence \cite{wachsmuth1994recognition}) and proposed simple yet effective update rules.
Hoyer \cite{hoyer2002non} combined sparse coding and NMF that allows control over sparseness.
Ding investigated the equivalence between the NMF and K-means clustering in \cite{ding2005equivalence,ding2006orthogonal} and presented simple update rules for orthogonal NMF.
Wang \cite{wang2013nonnegative} provided a comprehensive review on recent processes in the NMF area.

\subsection{Network Architecture, Training Strategy and Verification}

\begin{figure}[t]
\centering
\includegraphics[width=3.3in]{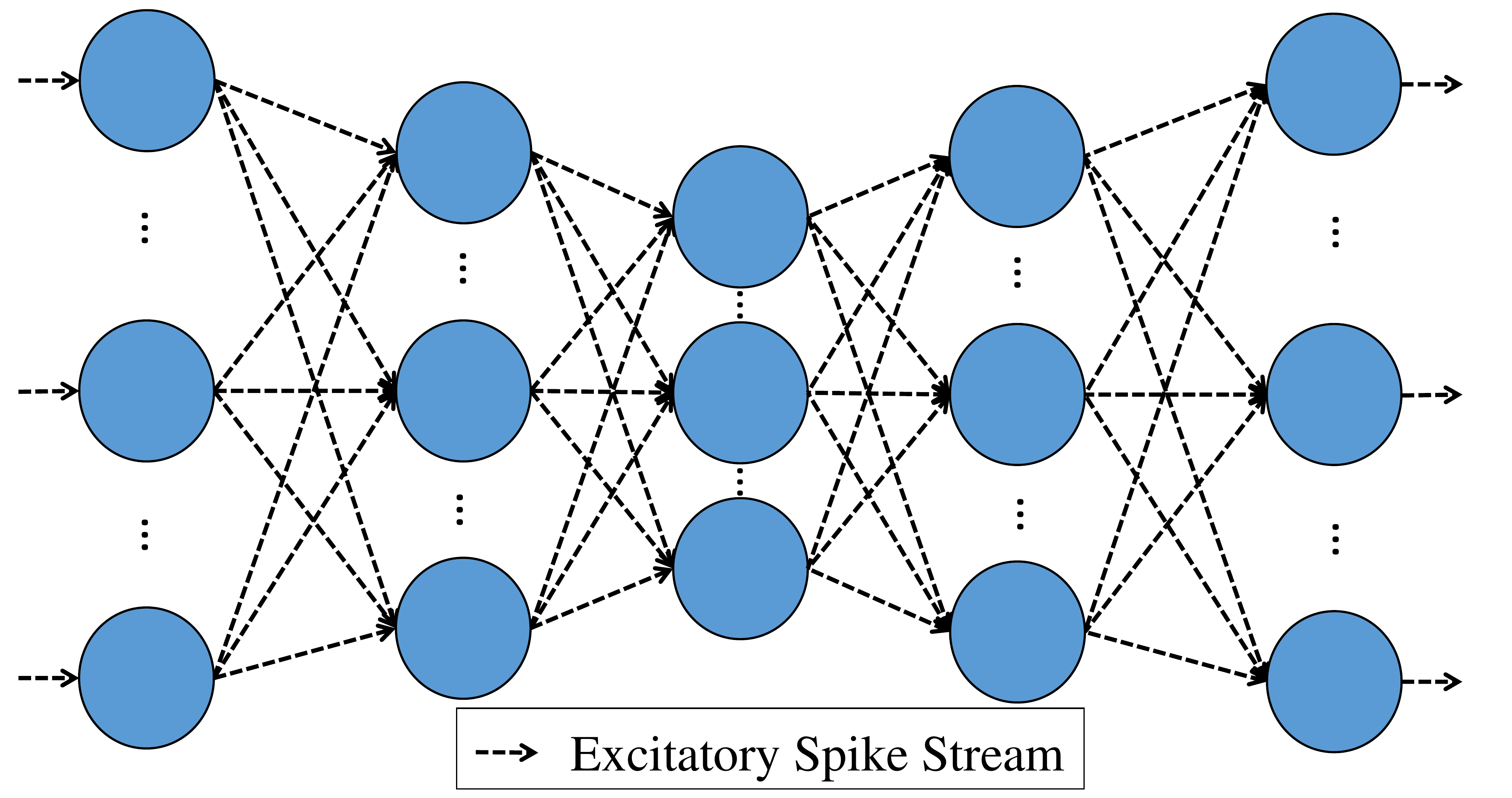}
\caption{Structure of a multi-layer non-negative RNN autoencoder.}
\label{fig.LRNN_multi_layer}
\end{figure}

In \cite{2016arXiv160908151Y} and Chapter 4 of \cite{phd-yonghua2018}, the authors first exploit the structure of the RNN equations as a quasi-linear structure. Using it in the feed-forward case, an RNN-based shallow non-negative autoencoder is constructed. This shallow autoencoder is then stacked into a multi-layer feed-forward autoencoder following the network architecture in the deep learning area \cite{hinton2006reducing,hinton2006fast,lecun2015deep}. The structure of the resultant multi-layer non-negative RNN autoencoder is given in Figure \ref{fig.LRNN_multi_layer}.

Since connecting weights in the RNN are products of firing rates and transition probabilities, they are subject to the constraints of non-negativity and that the sum of probabilities is no larger than 1, which are called the RNN constraints.
In view of this, the conventional gradient descent is not applicable for training such an autoencoder.
By adapting the update rules from non-negative graph embedding \cite{NNPCA} that
is closely related to
NMF, applicable update rules are developed for the autoencoder, which satisfy the first RNN constraint of non-negativity. For the second RNN constraint, a check-and-adjust procedure is imposed into the iterative learning process of the learning algorithms. The training procedure of the SGD is also adapted into the algorithms.

\begin{figure}[t]\centering
\subfigure[MNIST]{\label{fig.mnist_multi_layer}\includegraphics[width=2in]{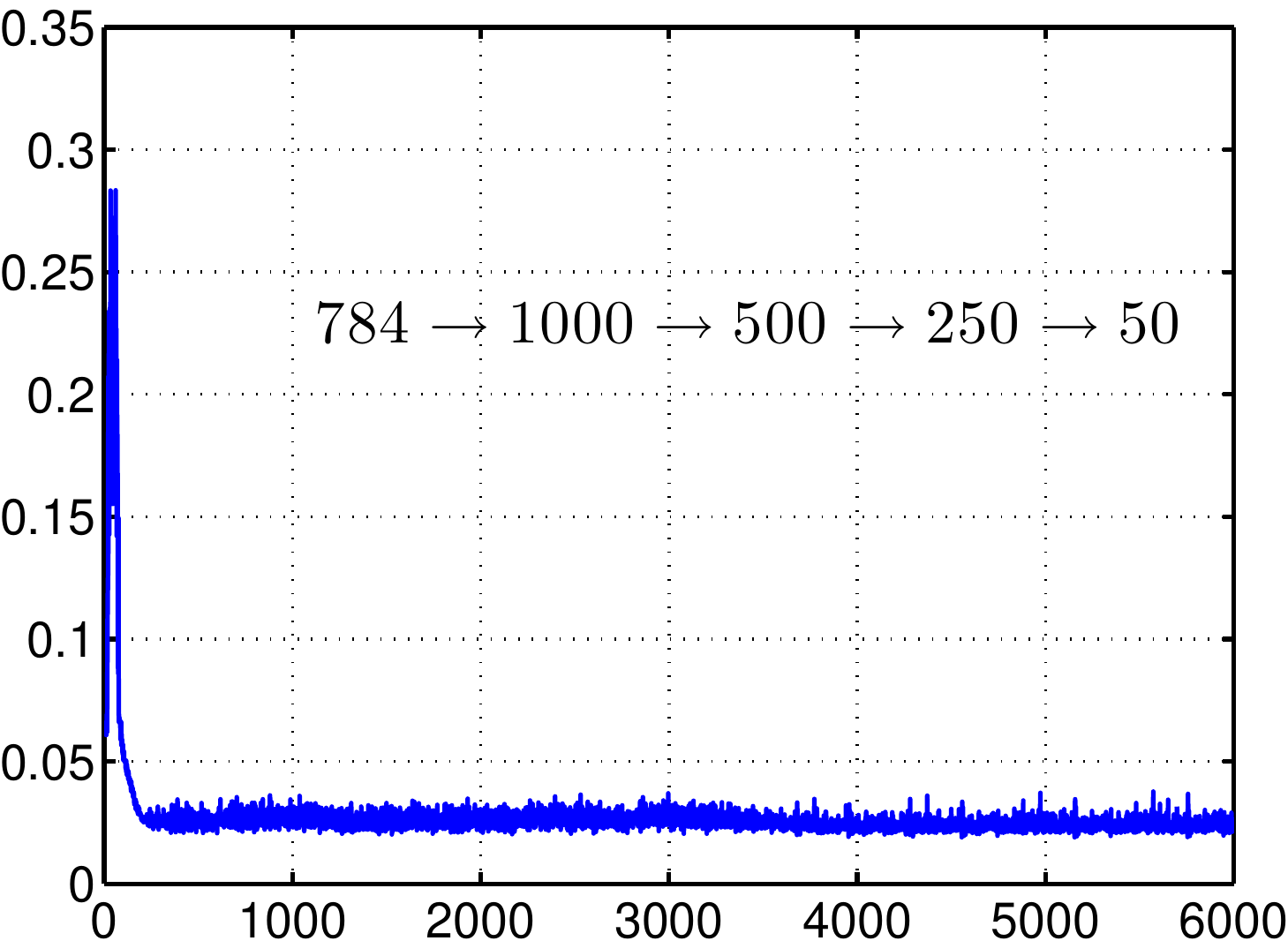}}
\subfigure[Yale face]{\label{fig.yale_multi_layer}\includegraphics[width=2in]{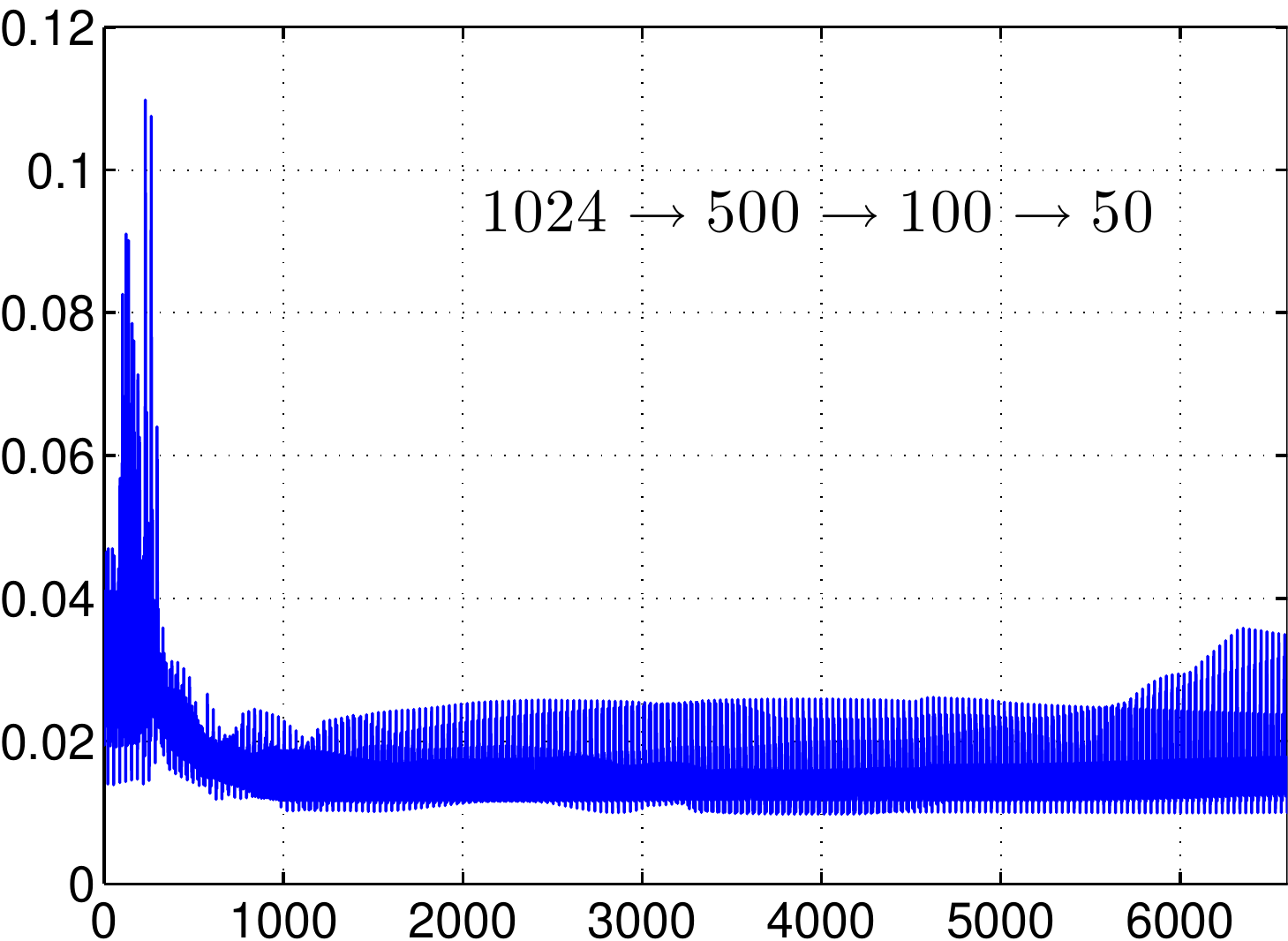}}
\subfigure[CIFAR-10]{\label{fig.cifar_10_multi_layer}\includegraphics[width=2in]{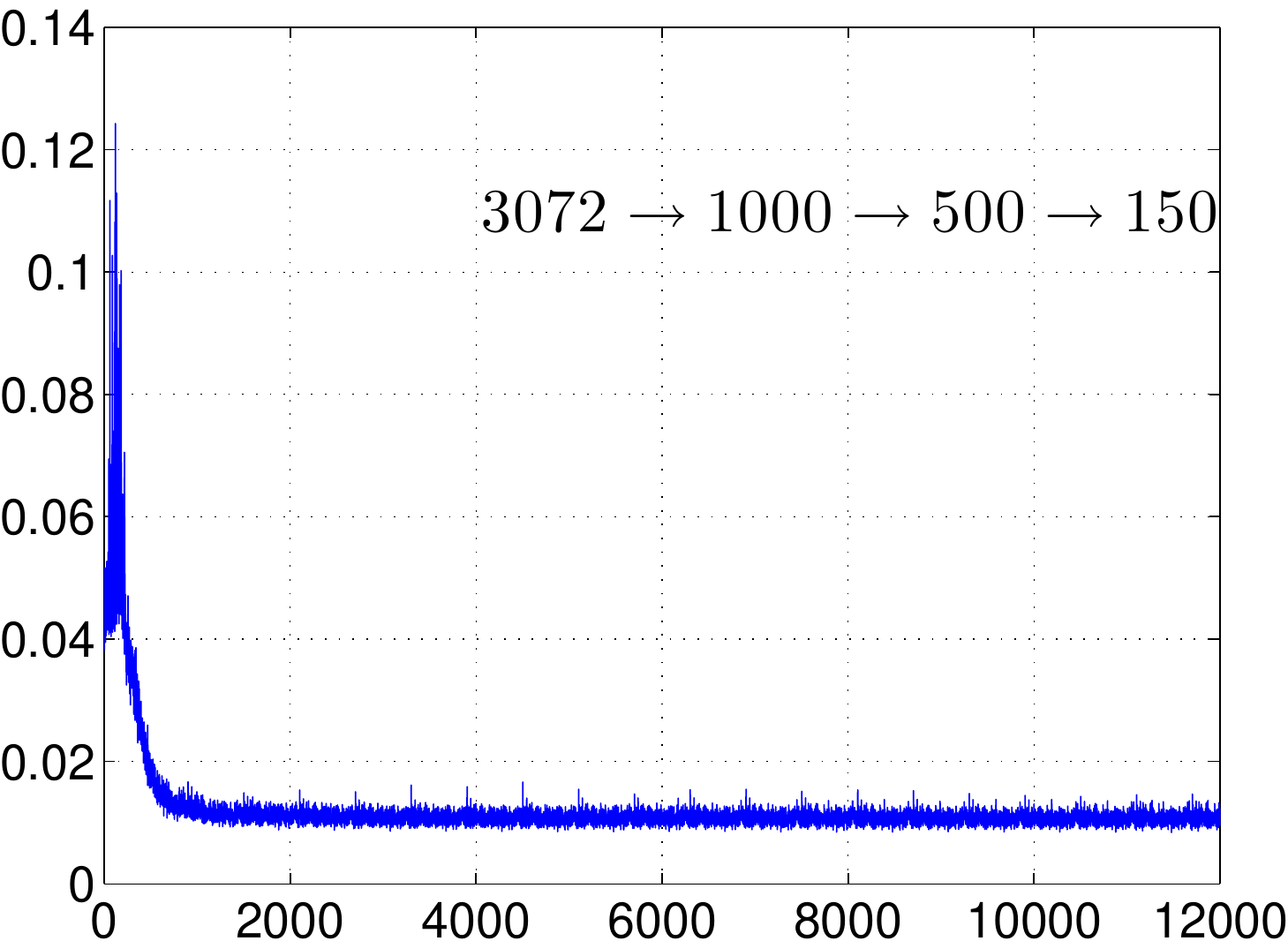}}
\caption{Reconstruction error (Y-axis) versus iteration number (X-axis) in training multi-layer RNN autoencoders for the MNIST, Yale face and CIFAR-10 datasets.} \label{fig.mnist_yale_cifar}
\end{figure}

Figure \ref{fig.mnist_yale_cifar} presents the curves of reconstruction error versus iteration number by using multi-layer RNN autoencoders to learn the MNIST \cite{mnist}, Yale face \cite{cai2007learning} and CIFAR-10 \cite{krizhevsky2009learning} datesets. It demonstrates the convergence of proposed RNN autoencoders. In \cite{2016arXiv160908151Y} and Chapter 4 of \cite{phd-yonghua2018}, the learning efficacy of the non-negative autoencoders equipped with the learning algorithms is well verified via numerical experiments on both typical image datasets including the MNIST, Yale face and CIFAR-10 datesets and 16 real-world datasets in different areas from the UCI machine learning repository \cite{Lichman:2013}, in terms of convergence and reconstruction performance and dimensionality-reduction capability for pattern classification.

\section{Deep Learning with Dense Random Neural Network}\label{chapter.densernn}

The work by the authors in \cite{gelenbedeep2016,gelenbedeep2016_SAI,yin2016deep,gelenbe2017deep} and Chapter 5 of \cite{phd-yonghua2018} explores the idea that the human brain contains many important areas that are composed of dense clusters of cells, such as the basal ganglia and various nuclei. These clusters may be composed of similar or identical cells, or of cells of different varieties; however, due to the density of their arrangement, it may be that such clusters allow for a substantial or at least increased amount of direct communication between somata, in addition to the commonly exploited signalling through dendrites and synapses. Thus, based on recurrent RNN \cite{RNN89}, a mathematical model of dense clusters/nuclei is developed that models both synapses and direct soma-to-soma interactions, creating a dense random neural network (Dense RNN). Each cluster is modelled as a recurrent spiking RNN. Each neuron in each nucleus has a statistically identical interconnection structure with the other cells in the same nucleus; this statistical regularity allows for great individual variability among neurons both with regard to spiking times and interconnection patterns.
In the model of the dense nucleus, the authors define the number of cells as $N$, the probability for repeated firing as $p$, the firing rate as $r$, the external excitatory input as $\lambda^+$ and external inhibitory input as $\lambda^-$. Let $q$ denote the stationary excitation probability of a cell in the nucleus. Mathematical deduction has demonstrated that $q = \zeta(x,N,p,r,\lambda^+,\lambda^-)$, where $x$ is the external inputs from the cells outside the nucleus.
The explicit expression of $\zeta(\cdot)$ and details of the mathematical deduction can be found in Chapter 5 of \cite{phd-yonghua2018}. Theoretical analyses of the nucleus activation function $q = \zeta(x,N,p,r,\lambda^+,\lambda^-)$ are conducted to find the conditions of the values $N,p,r,\lambda^+,\lambda^-$ to maintain the stability of the nucleus. The conditions are summarized as $N \geq 2$, $0 \leq p \leq 1$, $\lambda^+>0,\lambda^->0,r>0$ and $\lambda^- \geq \lambda^+$.

\begin{figure}[t]
\centering
\includegraphics[width=2.8in]{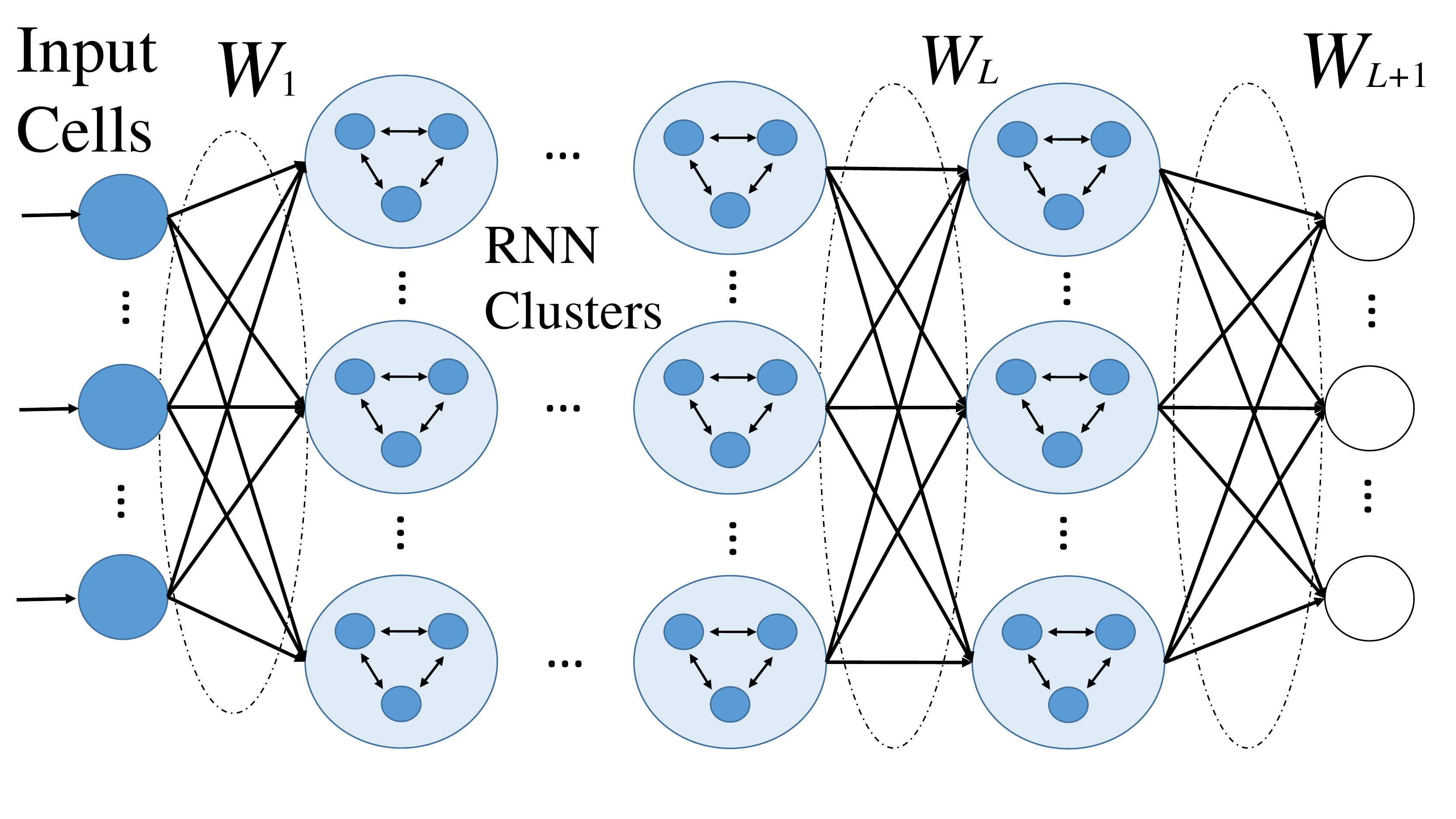}
\caption{Schematic representation of the MLDRNN.}
\label{fig.RNN_MLA}
\end{figure}

The model of dense clusters allows consideration of a multi-layer architecture (MLA) network wherein each layer is composed of a finite number of dense nuclei.
The MLA of dense nuclei, modelled by the Dense RNN and named as MLDRNN, is given schematically in Figure \ref{fig.RNN_MLA}.
Within each nucleus, the cells communicate with each other in fully-connected recurrent structures that use both synapse and direct soma-to-soma interaction \cite{Timotheou}. The communication structure between the various layers of nuclei is a conventional multi-layer feedforward structure: the nuclei in the first layer receive excitation signals from external sources, while each cell in each nucleus creates an inhibitory projection to the layer up.
Further, an efficient training procedure is developed for the MLDRNN, which combines unsupervised and supervised learning techniques.
Note that most of the work related to the training of the deep neural networks, as well as the RNN, uses the gradient-descent algorithms, where
there are inherent disadvantages of slow convergence and the possibilities of being trapped into
poor local minimums. The proposed training algorithm by the authors
converts the training of the MLDRNN into multiple convex optimization problems that
can be solved with comparable or better solutions in a much faster speed than the gradient-descent algorithms.
Specifically, let us denote the connecting weight matrices between layers of a $L$-hidden-layer MLDRNN by $W_l=[w_{h_{l},h_{l+1}}] \in \mathbb{R}_{\geq 0}^{H_{l} \times H_{l+1}}$ with $l=1,2,\cdots,L$.
In addition, let us define a readout matrix as $W_{L+1} \in \mathbb{R}^{H_{L+1} \times H_{L+2}}$, where $H_{L+2}$ is the dimension of the outputs.
The connection weights $W_l$ with $l=1,2,\cdots,L-1$ between the 1st and $L-1$ layers are determined by constructing a series of reconstruction optimization problems, subject to the nonnegative constraints of the connection weigths. These optimization problems are designed as convex problems and therefore the solving process can be very fast.
The connection weights $W_L$ is randomly generated within their feasible region. Finally, the determination of the matrix $W_{L+1}$ is formulated as a convex optimization problem that can be solved directly by the Moore-Penrose pseudo-inverse \cite{gelenbedeep2016,zhang2014cross,yin2012weights,zhang2012pruning,elm,mlelm}.

\begin{figure}[t]
\centering
\includegraphics[width=3.5in]{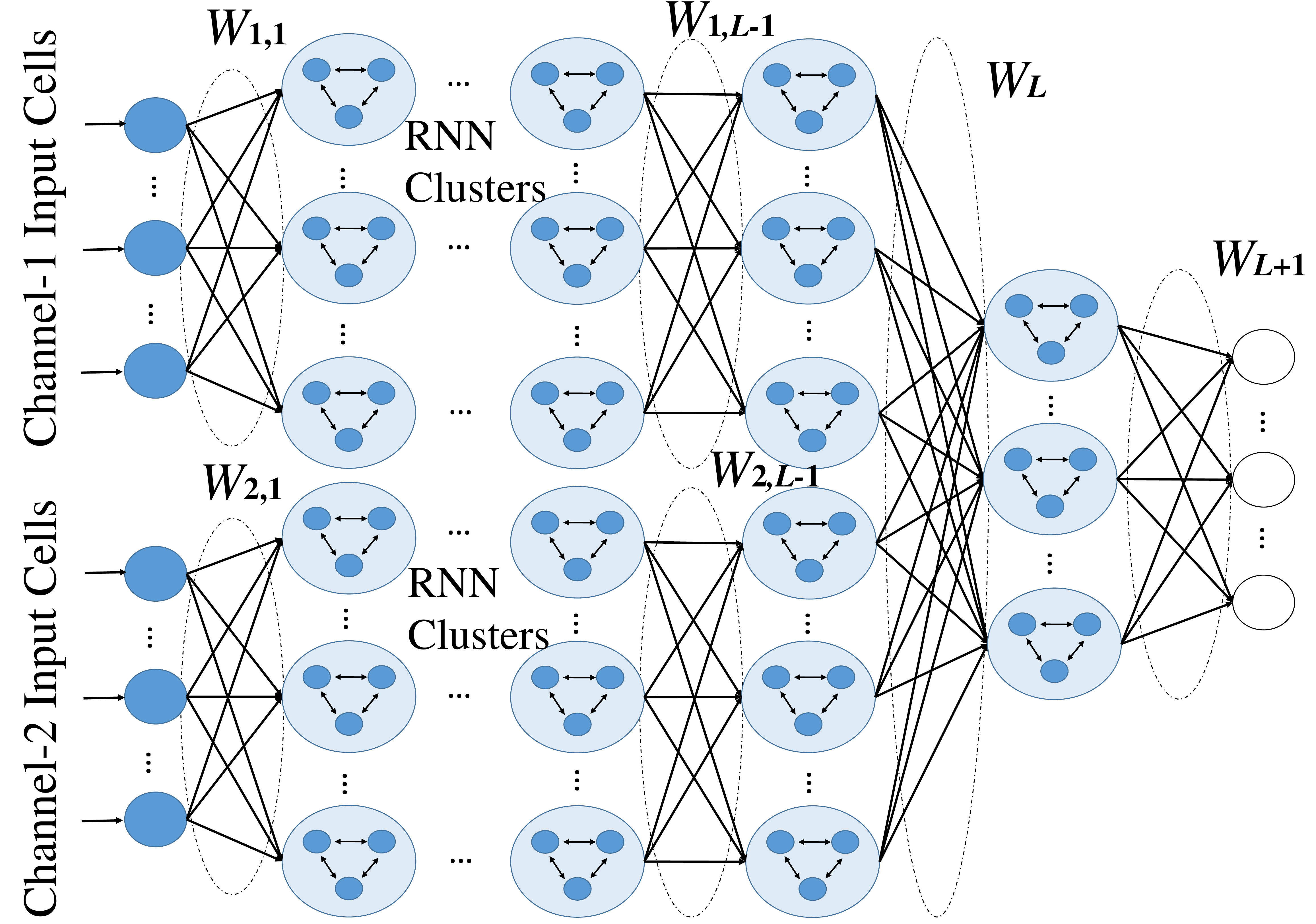}
\caption{Schematic representation of the MCMLDRNN.}
\label{fig.MCRNN}
\end{figure}

In the work of \cite{yin2016deep,gelenbe2017deep}, the multi-channel MLDRNN (MCMLDRNN), extended from the MLDRNN, is developed to handle multi-channel classification datasets. The schematic representation of the MCMLDRNN is given in Figure \ref{fig.MCRNN}.
The MCMLDRNN is applied to recognizing 3D objects, distinguishing different chemical gases and detecting daily and sports activities of humans. The corresponding numerical results on these three applications indicate that the MCMLDRNN is more accurate than state-of-the-art methods in most cases with high training efficiency, including the multi-layer perception (MLP), the convolutional neural network (CNN) and hierarchical extreme learning machine (H-ELM) \cite{mlelm}.
For example, Table \ref{tab.sport} presents the training and testing accuracies and time of different methods on the Daily and Sports Activities (DSA) dataset \cite{altun2010comparative,barshan2014recognizing,altun2010human} for detecting human activities.
From the table, it can be seen that the MCMLDRNN achieves the highest testing accuracy and the training time is much lower than the CNN with the dropout technique \cite{srivastava2014dropout} that achieves the second highest testing accuracy.
\begin{table*}[t]
\caption{Training and testing accuracies ($\%$) and time (s) of different methods on DSA dataset for detecting human activities.} \label{tab.sport}
\begin{center}
\begin{tabular}{|l|l|l|l|l|}
\hline
\multirow{2}*{Method}  &\multicolumn{2}{|l|}{Accuracies ($\%$)} &\multicolumn{2}{|l|}{Times (s)} \\\cline{2-5}
 &Training &Testing&Training &Testing\\	\hline
 MLP	&	99.61	$\pm	0.35	$	&	97.19	$\pm	0.58	$	&	307.87	$\pm	28.34	$	&	0.21	$\pm	0.01	$	\\	\hline
MLP+dropout	&	99.56	$\pm	0.37	$	&	97.47	$\pm	0.92	$	&	308.40	$\pm	15.02	$	&	0.31	$\pm	0.02	$	\\	\hline
CNN	&	99.84	$\pm	0.13	$	&	98.50	$\pm	0.38	$	&	1028.95	$\pm	3.38	$	&	1.31	$\pm	0.02	$	\\	\hline
CNN+dropout	&	99.84	$\pm	0.12	$	&	99.11	$\pm	0.20	$	&	1047.59	$\pm	1.72	$	&	1.38	$\pm	0.02	$	\\	\hline
MLP*	&	99.00	$\pm	0.95	$	&	96.68	$\pm	1.32	$	&	99.86	$\pm	24.28	$	&	0.26	$\pm	0.12	$	\\	\hline
MLP+dropout*	&	99.69	$\pm	0.26	$	&	97.62	$\pm	0.67	$	&	82.38	$\pm	2.04	$	&	0.16	$\pm	0.01	$	\\	\hline
CNN*	&	99.78	$\pm	0.21	$	&	98.54	$\pm	0.44	$	&	479.48	$\pm	421.34	^\dag$	&	0.17	$\pm	0.04	$	\\	\hline
CNN+dropout*	&	99.87	$\pm	0.1	$	&	98.96	$\pm	0.25	$	&	65.79	$\pm	2.51	$	&	0.17	$\pm	0.02	$	\\	\hline
H-ELM	&	98.08	$\pm	0.65	$	&	96.83	$\pm	0.61	$	&	6.55	$\pm	0.34	$	&	0.66	$\pm	0.11	$	\\	\hline
Original MLDRNN \cite{gelenbedeep2016}	&	98.41	$\pm	0.60	$	&	91.43	$\pm	1.60	$	&	4.55	$\pm	0.13	$	&	0.41	$\pm	0.03	$	\\	\hline
Improved MLDRNN	\cite{gelenbe2017deep} &	94.68	$\pm	0.65	$	&	91.74	$\pm	0.46	$	&	10.54	$\pm	0.13	$	&	0.62	$\pm	0.04	$	\\	\hline
MCMLDRNN	&	99.79	$\pm	0.03	$	&	\textbf{99.13}	$\pm	0.18	$	&	17.40	$\pm	0.41	$	&	3.08	$\pm	0.29	$	\\	\hline
MCMLDRNN1	\cite{gelenbe2017deep}&	99.59	$\pm	0.05	$	&	98.88	$\pm	0.13	$	&	40.90	$\pm	0.43	$	&	7.18	$\pm	0.26	$	\\	\hline
MCMLDRNN2	\cite{gelenbe2017deep}&	98.51	$\pm	0.14	$	&	94.41	$\pm	0.39	$	&	114.60	$\pm	3.84	$	&	5.66	$\pm	0.38	$	\\	\hline
\multicolumn{5}{l}{*These experiments are conducted using the GPU.}\\
\multicolumn{5}{l}{$^\dag$ The data is 65.31$\pm 7.17$(s) if the experiment with 900.28s (unknown error) is removed.}
\end{tabular}
\end{center}
\end{table*}
More related results can be found in \cite{yin2016deep,gelenbe2017deep} and Chapter 5 of \cite{phd-yonghua2018}.
Note that, in addition to the above applications, the work in \cite{yin_cloud2017} has successfully applied the dense RNN to inferring the states of servers within the Cloud system.

\begin{table}[t]
\caption{Results by using dense nuclei and very dense nuclei in 100 trials to classify DSA dataset.} \label{tab.compare_dsa}
\begin{center}
\begin{tabular}{|l|l|l|l|l|}
\hline
 &Dense Nuclei &Very Dense Nuclei\\	\hline
Training accuracy ($\%$) &99.79$\pm 0.05$ &99.78$\pm 0.04$ \\	\hline
Testing accuracy ($\%$) &99.13$\pm 0.18$ &99.16$\pm 0.15$\\	\hline
Training time (s)&21.38$\pm 0.45$ &13.10$\pm 0.35$\\	\hline
Testing time (s)&3.39$\pm 0.15$ &1.14$\pm 0.08$\\	\hline
\end{tabular}
\end{center}
\end{table}

Finally, in Chapter 5 of \cite{phd-yonghua2018}, the case where the number of cells in a nucleus is very large is investigated, by which the transfer functions of dense nuclei are significantly simplified.
Table \ref{tab.compare_dsa} presents the results by using dense nuclei and very dense nuclei in 100 trials to classify DSA dataset. It can be seen that both the training and testing times have been reduced by using the very dense nuclei. More related results can be found in Chapter 5 of \cite{phd-yonghua2018}.
The numerical results show that a dense-enough nucleus produces a more efficient algorithm than a not-that-dense nucleus.

\section{Deep Learning with Standard Random Neural Network} \label{chapter.mlrnn}

The authors' recent work in \cite{yyh_rnnl_ijcnn2017} and Chapter 6 of \cite{phd-yonghua2018} goes back to the original simpler structure of the RNN and investigates the power of single standard RNN cells for deep learning in the two aspects.  This section presents more detailed and sharper descriptions and procedures.

In the first part, it is shown that, with only positive parameters, the RNN implements image convolution operations similar to the convolutional neural networks \cite{mnist,lecun2015deep,springenberg2014striving}.
The authors examine single-cell, twin-cell and cluster approaches that model the positive and negative parts of a convolution kernel as the arrival rates of excitatory and inhibitory spikes to receptive cells. The relationship between the RNN and ReLU activation \cite{AISTATS2011_GlorotBB11} is also investigated and an approximate ReLU-activated convolution operation is presented.

In the second part, a multi-layer architecture of the standard RNN (MLRNN) is built.
The computational complexity of the MLRNN is lower than the MLDRNN model developed in \cite{yin2016deep,gelenbe2017deep}, and it can also be generalized to handle multi-channel datasets.
Numerical results in \cite{yyh_rnnl_ijcnn2017} and Chapter 6 of \cite{phd-yonghua2018} regarding multi-channel classification datasets show that the MLRNN is effective and that it is arguably the most efficient among five different deep-learning approaches.

\subsection{Image Convolution with Single-Cell Approach}

Suppose a single RNN cell receives excitatory and inhibitory Poisson spike trains from other cells with rates $x^{+}$ and $x^{-}$, respectively, and it also receives an excitatory Poisson spike train from the outside world with rate $\lambda^+$. The firing rate of the cell is $r$. Then, based on the RNN theory \cite{RNN89,RNN90,RNN93} and RNN equation (\ref{eqn.rnn}), the probability in the steady state that this cell is excited can be calculated by
\begin{equation}
q=\min(\frac{\lambda^++ x^{+} }{r + x^{-}},1). \label{RNN_cell}
\end{equation}
For notational ease, let $\phi(x^{+},x^{-})|_{\lambda,r}=\min(({\lambda+x^{+}})/({r+x^{-}}),1)$ and
$\phi(\cdot)$ is used as a term-by-term function for vectors and matrices.

The convolution operation of this approach with single cells is shown schematically in Figure \ref{fig.single_cell}.
As see from the figure, the quasi-linear RNN cells (called the LRNN-E cell) \cite{2016arXiv160908151Y} presented in Section \ref{chapter.lrnn} are utilized as input cells for the convolution inputs $I$. For these input cells, the firing rates are set as 1, the rates of the external inhibitory spikes are set as 0 and the rates of the external excitatory spikes are set as $I$. By assuming that $0 \leq I \leq 1$, the stationary excitation probabilities of these input cells are $I$.

Second, the convolution kernel denoted by the kernel weight matrix $W$ is normalized to satisfy the RNN probability constraint via $W \leftarrow W/(\text{sum}(|W|))$, where the operation sum$(\cdot)$ produces the summation of all input elements. Then, $W^{+}=\max(W,0) \geq 0$ and $W^{-}=\max(-W,0) \geq 0$ to avoid negativity,
where term-by-term operation $\max(a,b)$ produces the larger element between $a$ and $b$. As shown in Figure \ref{fig.single_cell}, the input cells are connected to a set of RNN cells (\ref{RNN_cell}) in a local connectivity pattern (each RNN cell (\ref{RNN_cell}) is connected to only a small region of the input cells \cite{wiki_cnn}) and may fire excitatory and inhibitory spikes towards them.
The weights $W^{+}$ are utilized as the rates of excitatory spikes from the input cells to the RNN cells (\ref{RNN_cell}), while the weights $W^{-}$ are utilized as the rates of inhibitory spikes.
In this case, the stationary excitation probabilities of the RNN cells (\ref{RNN_cell}) denoted by a matrix $O$ can be obtained as
\begin{equation}
O=\phi(\text{conv}(I,W^{+}),\text{conv}(I,W^{-})),  \label{RNN_convolution_single1}
\end{equation}
where $\text{conv}(X,W)$ denotes a standard image convolution operation of input $X$ with the convolutional kernel  $W$. Or, in the other case, when $W^{-}$ and $W^{+}$ are utilized as excitatory and inhibitory spike rates, the expression of $O$ is
\begin{equation}
O=\phi(\text{conv}(I,W^{-}),\text{conv}(I,W^{+})).  \label{RNN_convolution_single2}
\end{equation}

\subsection{Image Convolution with Twin-Cell Approach}

\begin{figure}[t]\centering
\subfigure[Single-cell approach]{\label{fig.single_cell}\includegraphics[width=2.5in]{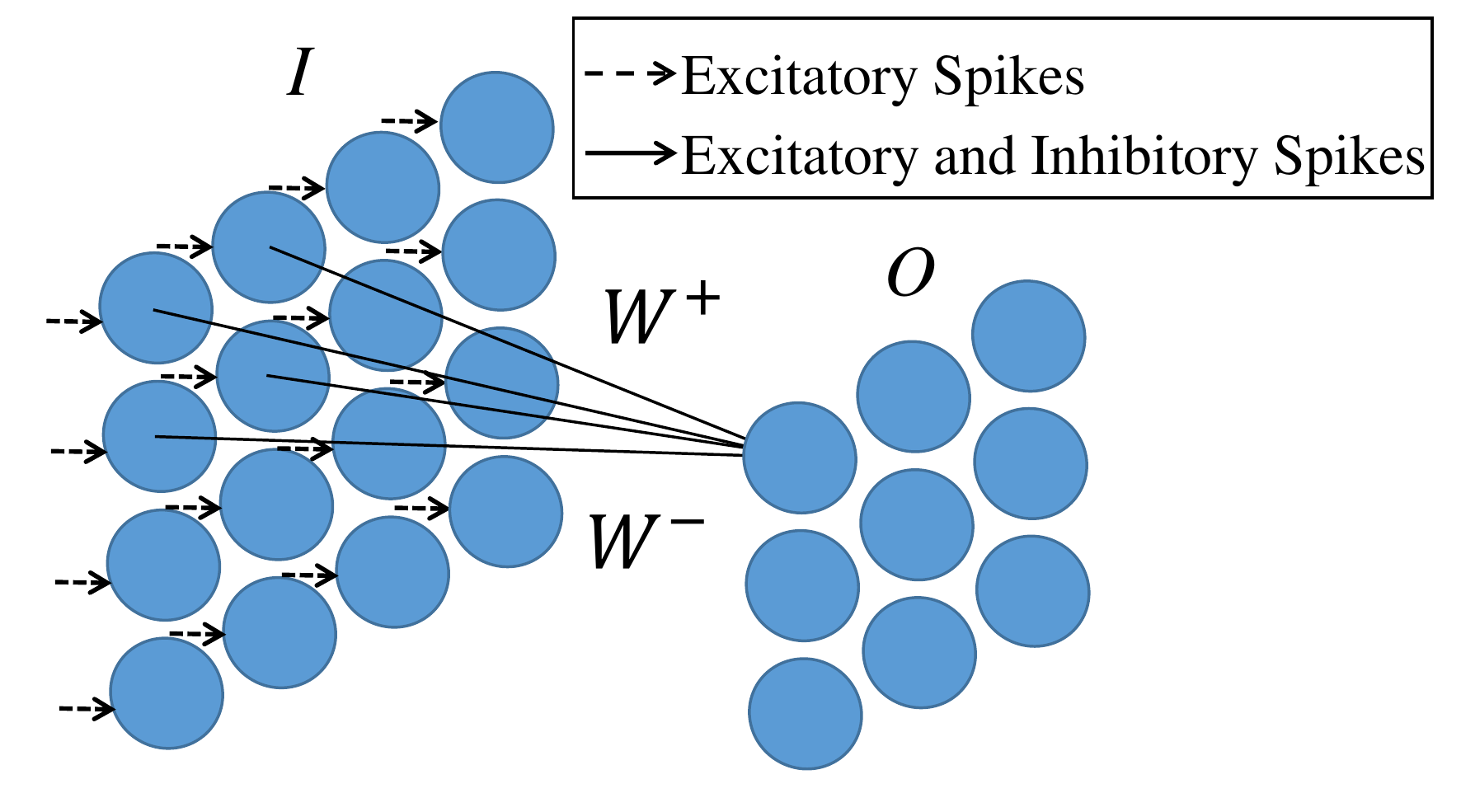}}
\subfigure[Twin-cell approach]{\label{fig.twin_cell}\includegraphics[width=2.5in]{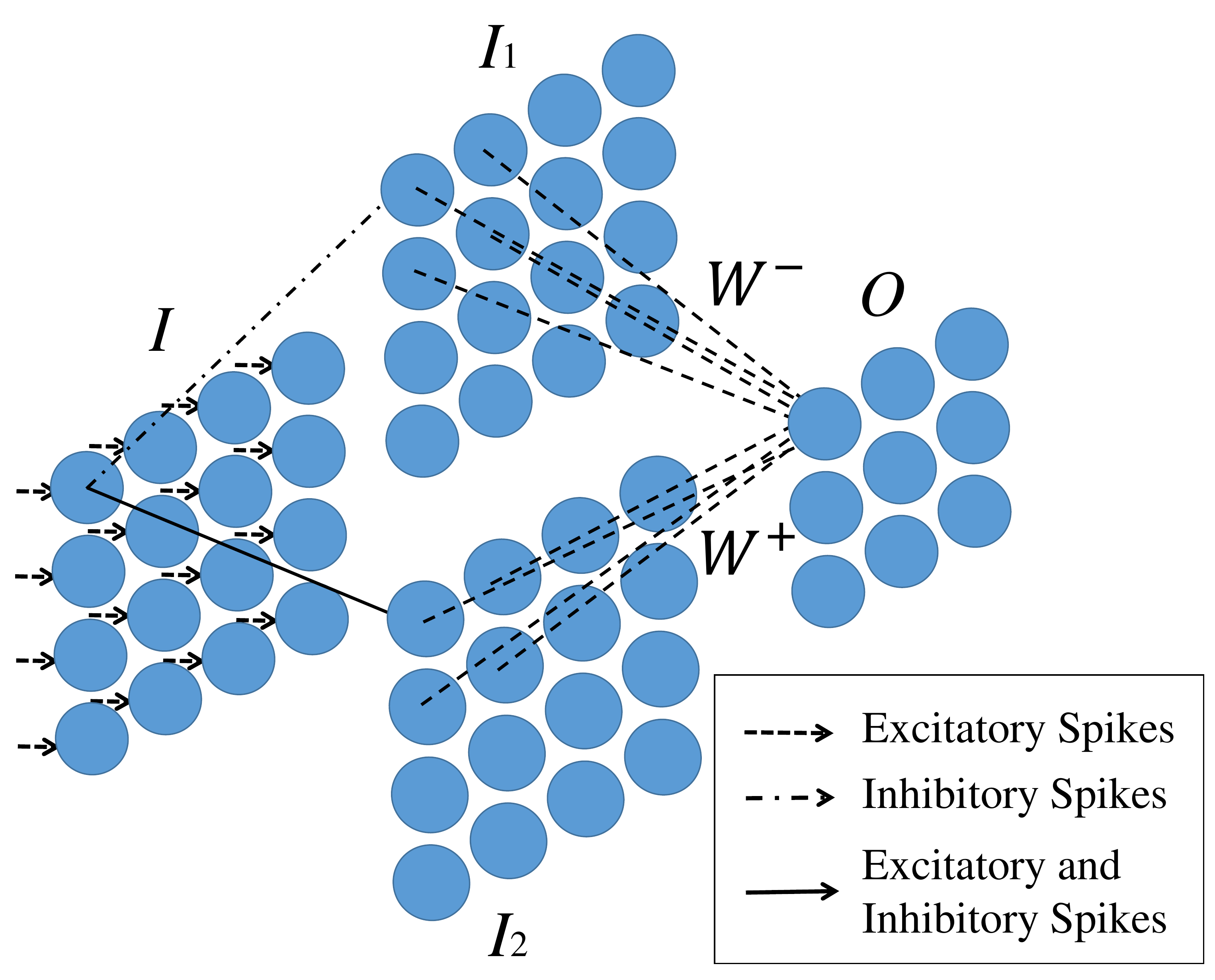}}
\caption{Image convolution operation with RNN using the single-cell and twin-cell approaches.}
\end{figure}

The convolution operation of this approach with twin cells is shown schematically in Figure \ref{fig.twin_cell}.
The input cells receive external excitatory spikes with rates $I$ and produce $I$. The input cells are then connected with twin arrays constituted of RNN cells (\ref{RNN_cell}). These two arrays have the same number of cells (\ref{RNN_cell}): for the upper array, $\lambda^+=1,r=1$; while for the other array, $\lambda^+=0,r=1$.  As seen from Figure \ref{fig.twin_cell}, inputs $I$ pass through the twin arrays and produce 
\begin{equation} \label{RNN_convolution_twin_I1}
I_1=\phi(0,I)|_{\lambda^+=1,r=1} = \frac{\textbf{1}}{\textbf{1}+I} = \textbf{1} - \frac{I}{\textbf{1}+I},
\end{equation}
\begin{equation} \label{RNN_convolution_twin_I2}
I_2=\phi(I,I)|_{\lambda^+=0,r=1} = \frac{I}{\textbf{1}+I} = \textbf{1} - \frac{\textbf{1}}{\textbf{1}+I},
\end{equation}
where $\textbf{1}$ is an all-one matrix with appropriate dimensions and the division between matrices is element-wise.

Since the probabilities are non-negative and less than 1, the convolution kernel needs to be adjusted via $W \leftarrow W/\text{max}(|\text{conv}(I_1,W)|)$ if $\text{max}(|\text{conv}(I_1,W)|)>1$.
To satisfy the RNN probability constraints, the convolution kernel is normalized via $W \leftarrow W/\text{sum}(|W|)$
and split into $W^{+}=\max(W,0) \geq 0$ and $W^{-}=\max(-W,0) \geq 0$.
As seen from Figure \ref{fig.twin_cell}, the cells in the twin arrays are then connected to a set of receptive cells, which are quasi-linear cells presented in Section \ref{chapter.lrnn}.
For these receptive cells, the firing rates are set as 1 and they receive excitatory spikes from the outside world with rate $1-\text{sum}(W^{-})$.
The stationary excitation probabilities of the receptive cells are then obtained as
\begin{equation}\label{RNN_convolution_twin}
\begin{split}
&O=\min(\text{conv}(I_1,W^{+})+\text{conv}(I_2,W^{-})+\textbf{1}-\text{sum}(W^{-}),\textbf{1})\\
&=\min(\text{conv}(I_1,W^{+})+\text{conv}(\textbf{1},W^{-})-\text{conv}(I_1,W^{-})+\textbf{1}-\text{sum}(W^{-}),\textbf{1})\\
&=\min(\text{conv}(I_1,W)+\textbf{1},\textbf{1}).
\end{split}
\end{equation}

\subsection{Image Convolution with a Cluster Approach} \label{sec.conv_cluster}
This approach is based on a type of multi-cell cluster that approximates the rectified linear unit (ReLU), where the ReLU unit has been widely-used in the deep-learning area \cite{lecun2015deep,AISTATS2011_GlorotBB11}. The cluster is constituted by the LRNN-E cell \cite{2016arXiv160908151Y} and another different quasi-linear RNN cell that is deduced from the first-order approximation of the RNN equation (\ref{eqn.rnn}).

\subsubsection{First-Order Approximation of the RNN Equation}

The RNN formula (\ref{eqn.rnn}) lends itself to various approximations such as the first-order approximation.
Specifically, according to (\ref{eqn.rnn}),
\begin{equation*}
q_h=\frac{\lambda_h^++\sum_{v=1}^Nq_{v}r_vp^+_{vh}}{r_h+\lambda_h^-+\sum_{v=1}^Nq_{v}r_vp^-_{vh}}.
\end{equation*}
Let $\Gamma=\lambda_h^-+\sum_{v=1}^Nq_{v}r_vp^-_{vh}$. Then,
\begin{equation*}
q_h=\frac{\lambda_h^++\sum_{v=1}^Nq_{v}r_vp^+_{vh}}{r_h+\Gamma}.
\end{equation*}
The partial derivative of $q_h$ with respect to $\Gamma$ is
\begin{equation*}
\frac{\partial q_h}{\partial \Gamma}=-\frac{\lambda_h^++\sum_{v=1}^Nq_{v}r_vp^+_{vh}}{(r_h+\Gamma)^2}.
\end{equation*}
Based on Taylor's theorem \cite{wiki_taylor}, $q_h$ can be approximated in a neighbourhood $\Gamma=a$ as:
\begin{eqnarray*}
&&q_h \approx {q_h}\mid_{\Gamma=a} + \frac{\partial q_h}{\partial \Gamma}\mid_{\Gamma=a} (\Gamma - a)
=\frac{\lambda_h^++\sum_{v=1}^Nq_{v}r_vp^+_{vh}}{r_h+a} -\frac{\lambda_h^++\sum_{v=1}^Nq_{v}r_vp^+_{vh}}{(r_h+a)^2} (\Gamma - a).
\end{eqnarray*}
Suppose the value of $a$ satisfies $r_h >> a$, then
\begin{eqnarray*}
&&q_h \approx \frac{\lambda_h^++\sum_{v=1}^Nq_{v}r_vp^+_{vh}}{r_h}
-\frac{\lambda_h^++\sum_{v=1}^Nq_{v}r_vp^+_{vh}}{r_h^2} (\Gamma - a)
 \approx \frac{\lambda_h^++\sum_{v=1}^Nq_{v}r_vp^+_{vh}}{r_h} (1-\frac{\Gamma}{r_h}+ \frac{a}{r_h})\\
&& \approx \frac{\lambda_h^++\sum_{v=1}^Nq_{v}r_vp^+_{vh}}{r_h} (1-\frac{\Gamma}{r_h}).
\end{eqnarray*}
Then, the first order approximation of formula (\ref{eqn.rnn}) is obtained as:
\begin{eqnarray} \label{apRNN}
q_h \approx \frac{\lambda_h^++\sum_{v=1}^Nq_{v}r_vp^+_{vh}}{r_h} (1-\frac{\lambda_h^-+\sum_{v=1}^Nq_{v}r_vp^-_{vh}}{r_h}).
\end{eqnarray}

\subsubsection{Quasi-Linear RNN cell receiving inhibitory spikes}

Let $w^+_{vh} = r_vp^+_{vh}=0$, $w^-_{vh} = r_v p^-_{vh}$, $\lambda_h^+=1$, $\lambda_h^-=0$, $r_h=1$ and $r_v=1$. Note that these settings do not offend the condition $r_h>>\lambda_h^-+\sum_{v=1}^Nq_{v}r_vp^-_{vh}$, which becomes $1 >> \sum_{v=1}^N w^-_{vh} q_{v}$. Since $r_v=1$, then $\sum_{h=1}^{N} w^-_{vh} = \sum_{h=1}^{N} p^-_{vh} \leq 1$.
The first-order approximation (\ref{apRNN}) is rewritten as:
\begin{eqnarray}
&&q_h = \frac{1}{1+\sum_{v=1}^N w^-_{vh} q_{v}}\approx \min(1- \sum_{v=1}^N w^-_{vh} q_{v},1),
\label{apRNN3}
\end{eqnarray}
subjecting to $\sum_{h=1}^{N} w^-_{vh} \leq 1$ and $1 >> \sum_{v=1}^N w^-_{vh} q_{v}$. Since $q_{v} \leq 1$, then $\sum_{v=1}^N w^-_{vh} q_{v} \leq \sum_{v=1}^N w^-_{vh}$, and the condition $1 >> \sum_{v=1}^N w^-_{vh} q_{v}$ could be relaxed to $1 >> \sum_{v=1}^N w^-_{vh}$, which is independent of the cell states.
This simplified RNN cell receiving inhibitory spikes is quasi-linear, which is thus called a LRNN-I cell.

\subsubsection{Relationship between RNN and ReLU}

Based on the LRNN-E cell \cite{2016arXiv160908151Y} and LRNN-I cell (\ref{apRNN3}), the relationship between the RNN and the ReLU activation function \cite{lecun2015deep,AISTATS2011_GlorotBB11} is investigated; and it is shown that a cluster of the LRNN-I and LRNN-E cells produces approximately the ReLU activation.

\begin{figure}[t]\centering
\subfigure[ReLU unit]{\label{fig.relu}\includegraphics[width=1.6in]{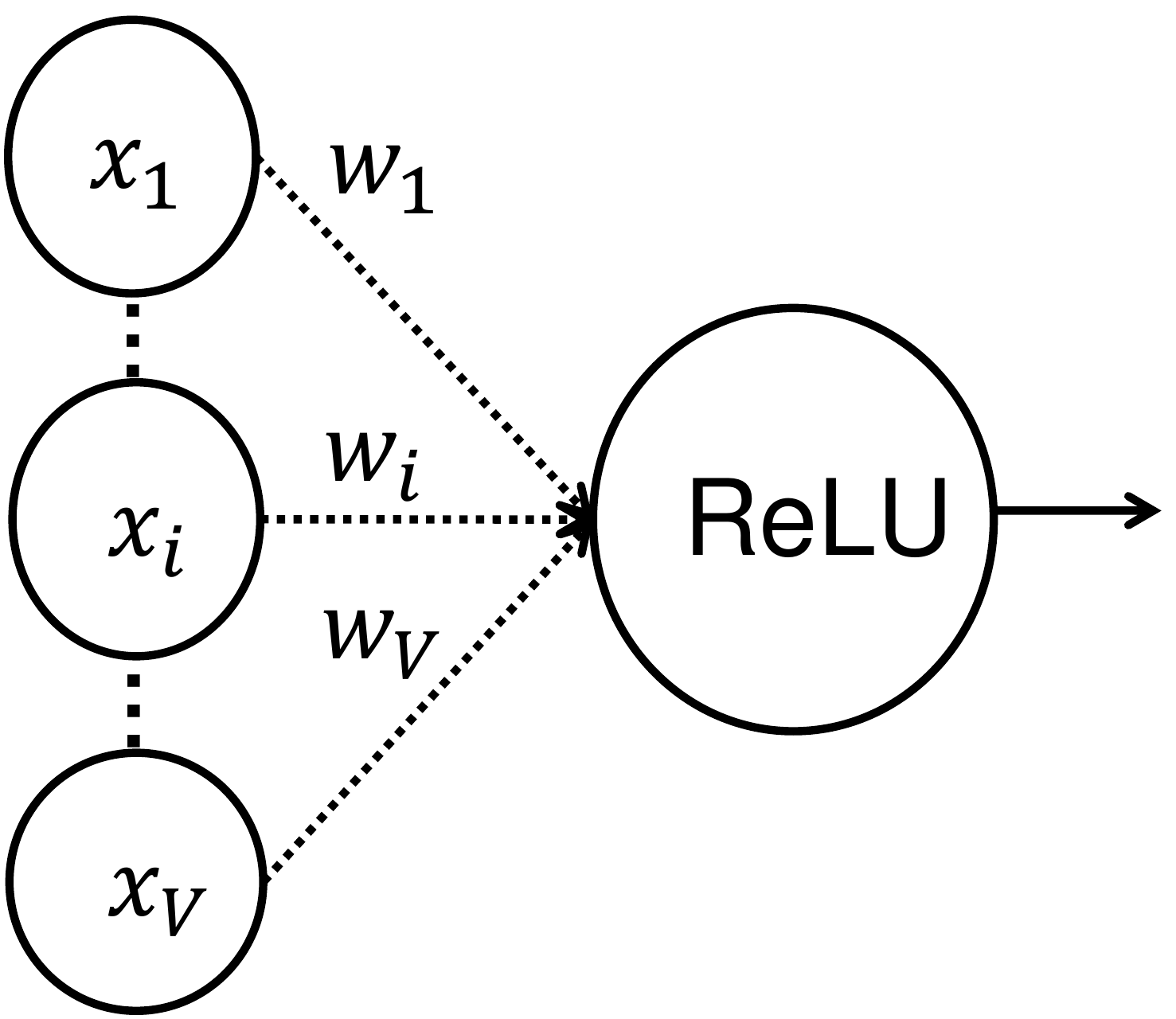}}
\subfigure[RNN cluster]{\label{fig.cluster_ReLU}\includegraphics[width=1.6in]{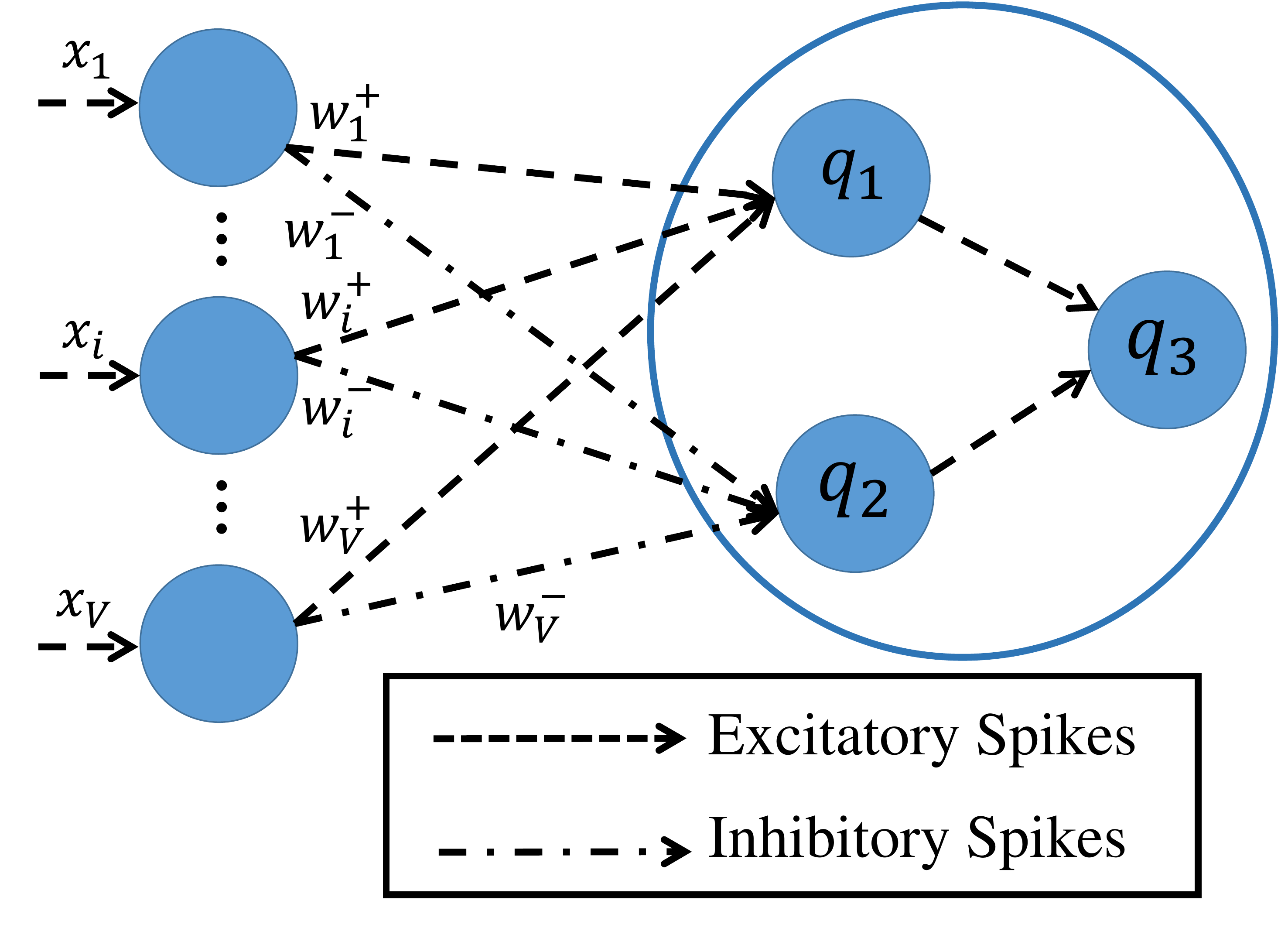}}
\caption{A ReLU unit and a cluster of the LRNN-I and LRNN-E cells to approximate the ReLU activation.}
\end{figure}

{\it ReLU Activation.}
A single unit with ReLU activation is considered. Suppose the input to the unit is a non-negative vector $X=[x_{1,v}] \in \mathbb{R}^{1 \times V}$ in the range $[0,1]$, and the connecting weights is a vector $W=[w_{v,1}] \in \mathbb{R}^{V \times 1}$ whose elements can be both positive and negative. Then, the output of this unit is described by ReLU$(XW)=\max(0,XW)$.
Figure \ref{fig.relu} illustrates the corresponding schematic representation of the ReLU.

{\it A Cluster of LRNN-I and LRNN-E Cells.}
A cluster of the LRNN-I and LRNN-E cells is capable of producing approximately the ReLU activation, whose schematic representation is given in
Figure \ref{fig.cluster_ReLU}. For illustration, let $W^{+}=[w^{+}_{v,1}]=\max(0,W)$ and $W^{-}=[w^{-}_{v,1}]=-\min(0,W)$. It is evident that $W^{+} \geq 0$, $W^{-} \geq 0$, $W=W^{+}-W^{-}$ and ReLU$(XW)=\max(0,XW^{+}-XW^{-})$.

First, the input $X$ is imported into
a LRNN-E cell (the 1st cell) with connecting weights $W^{-}$. Then the stationary excitation probability of the 1st cell is
\begin{equation*}
q_1 = \min(X W^{-},1).
\end{equation*}
Let us assume $X W^{-} \leq 1$. Then, $q_1 = X W^{-}$.

The input $X$ is also imported into a LRNN-I cell (the 2nd cell) with $W^{+}$, and its cell excitation probability is
\begin{equation*}
q_2 = \frac{1}{1+X W^{+}} \approx \min(1-X W^{+},1).
\end{equation*}
Based on (\ref{apRNN3}), the condition of this approximation is $\sum_{v=1}^V w^+_{v,1} < < 1$.
Suppose this condition holds, then $q_2 \approx 1 - X W^{+}$.

Second, the 1st and 2nd cells of $q_1$ and $q_2$ are connected to a LRNN-E cell (i.e., the 3rd cell or the output cell) with connecting weight being 1. The stationary excitation probability of the 3rd/output cell is
\begin{eqnarray*}
&&q_3 =  \min(q_2 + q_1,1) = \min(\frac{1}{1+ X W^{+}} + X W^{-},1)\approx \min(1- X W^{+} + X W^{-},1).
\end{eqnarray*}
In $q_3$, the information where $1-X W^{+} + X W^{-} = 1 - XW > 1$ (i.e., $XW < 0$) is removed by the LRNN-E cell. Then,
\begin{equation} \label{eqn.cluster}
q_3 =\varphi(X W^{+},X W^{-})  \approx 1-\text{ReLU}(XW),
\end{equation}
where, for notation ease, the activation of this cluster is defined as $\varphi(x^{+},x^{-})= \min({1}/({1+ x^{+}}) + x^{-},1)$ and use
$\varphi(\cdot)$ as a term-by-term function for vectors and matrices.
The conditions for the approximation in (\ref{eqn.cluster}) are $X W^{-} \leq 1$ (that can be loosed as $\sum_{v=1}^V w^-_{v,1} = \text{sum}(W^{-}) \leq 1$ if $X \leq 1$) and $\sum_{v=1}^V w^+_{v,1} = \text{sum}(W^{+}) << 1$.

\begin{figure}[t]
\centering
\includegraphics[width=3in]{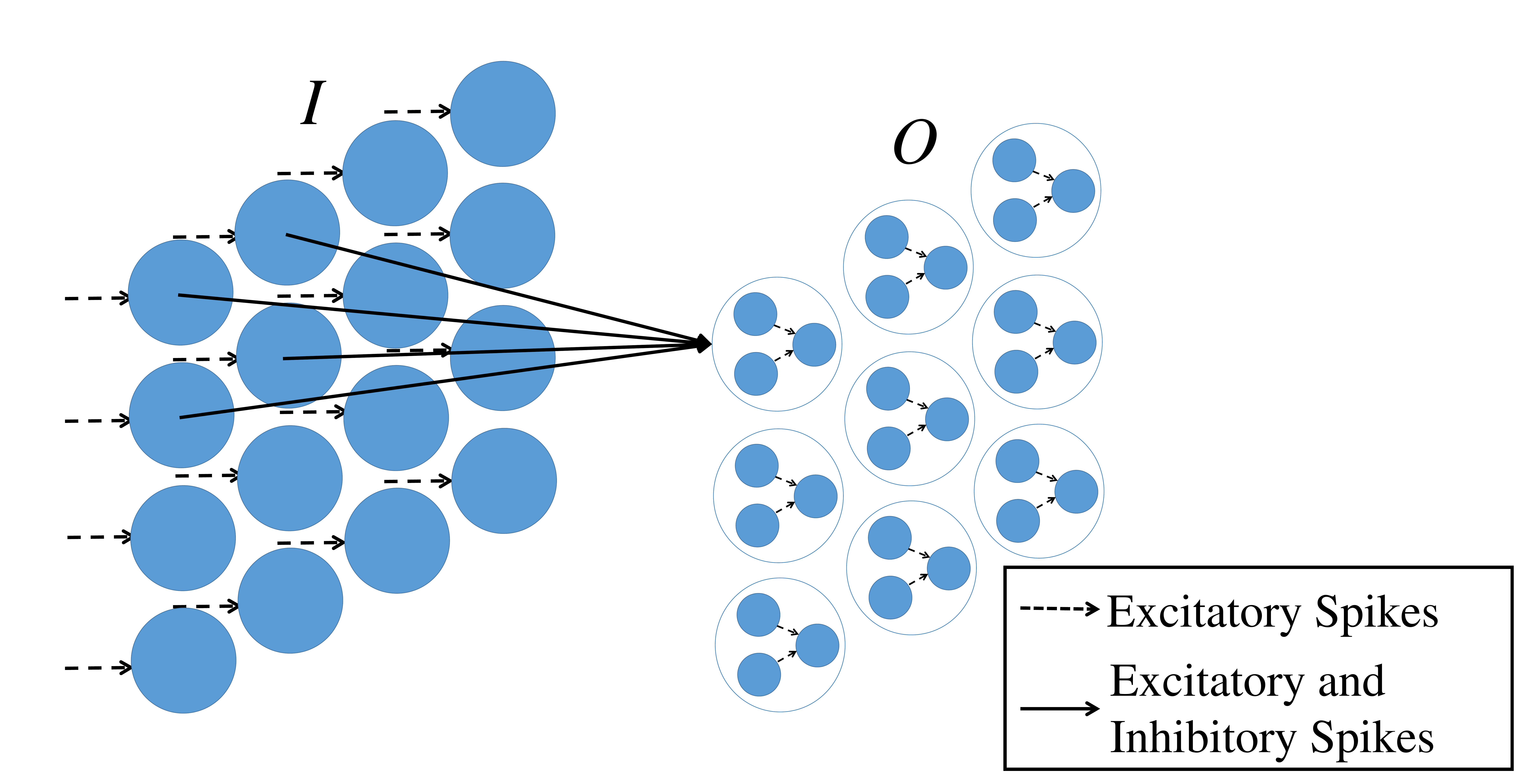}
\caption{An RNN convolution operation with the cluster approach.}
\label{fig.cluster}
\end{figure}

\subsubsection{Cluster-Based Convolution}

The convolution operation with clusters (\ref{eqn.cluster}) is shown schematically in Figure \ref{fig.cluster}.
As seen from the figure, the input cells receive excitatory spikes with rates $I$ and produce stationary excitation probabilities $I$.
Then, input cells are connected to a set of receptive clusters (\ref{eqn.cluster}) with local connectivity.
To satisfy the conditions for the approximation in (\ref{eqn.cluster}), the convolution kernel is normalized via $W \leftarrow W/\text{sum}(|W|)/10$.
In addition, let us split the kernel as $W^{+}=\max(W,0) \geq 0$ and $W^{-}=\max(-W,0) \geq 0$. By assuming that $0.1 << 1$, it is evident that $\text{sum}(W^{+})\leq 0.1 << 1$ and $\text{sum}(W^{-})\leq 0.1 << 1 $, which satisfy the approximation conditions.
Then, the weights $W^{+}$ and $W^{-}$ are utilized respectively as the rates of the excitatory and inhibitory spikes from the input cells to the clusters. The stationary excitation probabilities of the output cells of the receptive clusters are obtained as
\begin{eqnarray}\label{RNN_convolution_cluster1}
&&O=\varphi(\text{conv}(I,W^{+}),\text{conv}(I,W^{-})) \approx \textbf{1}-\text{ReLU}(\text{conv}(I,W)).
\end{eqnarray}

\begin{figure}[t]
\centering
\includegraphics[width=2in]{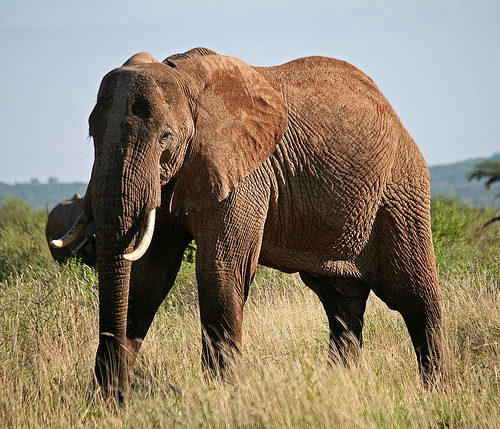}
\caption{The original image.}
\label{fig.original_image}
\end{figure}

\begin{figure}[t]\centering
\subfigure[ReLU activation]{\label{fig.ReLU_image}\includegraphics[width=3in,height=2.5in]{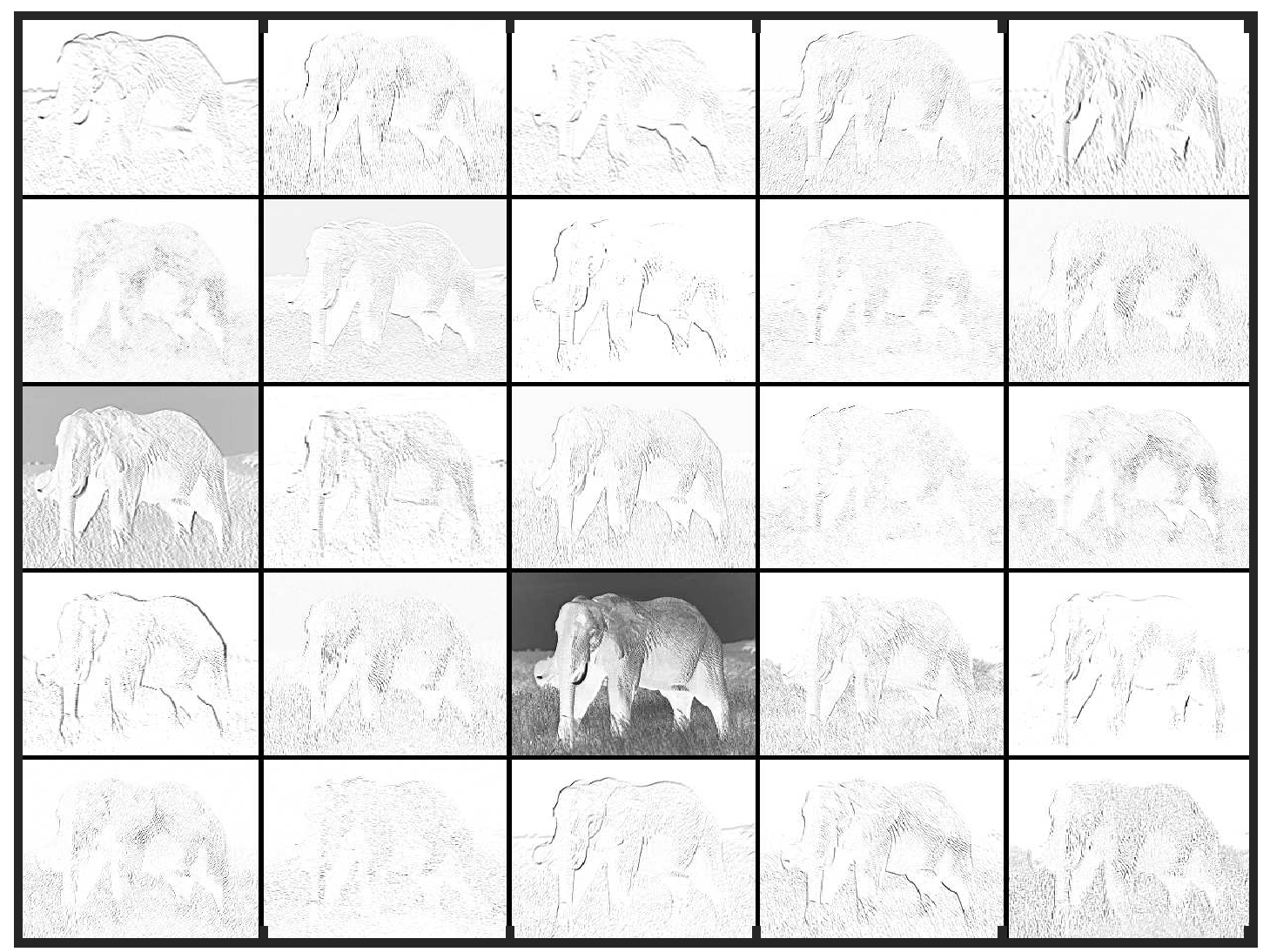}}
\subfigure[Single-cell approach]{\label{fig.single_cell_image}\includegraphics[width=3in,height=2.5in]{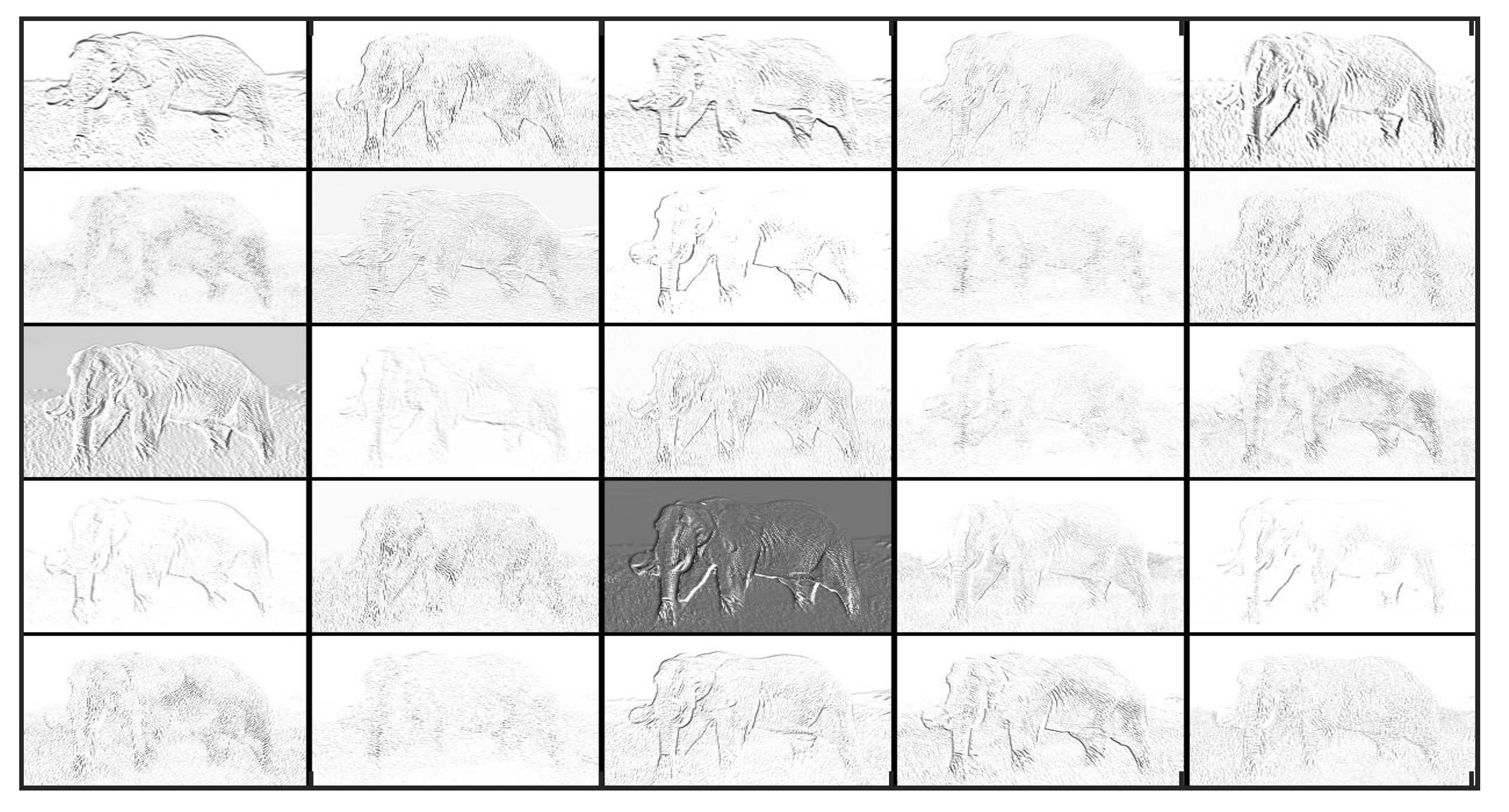}}
\subfigure[Twin-cell approach]{\label{fig.twin_cell_image}\includegraphics[width=3in,height=2.5in]{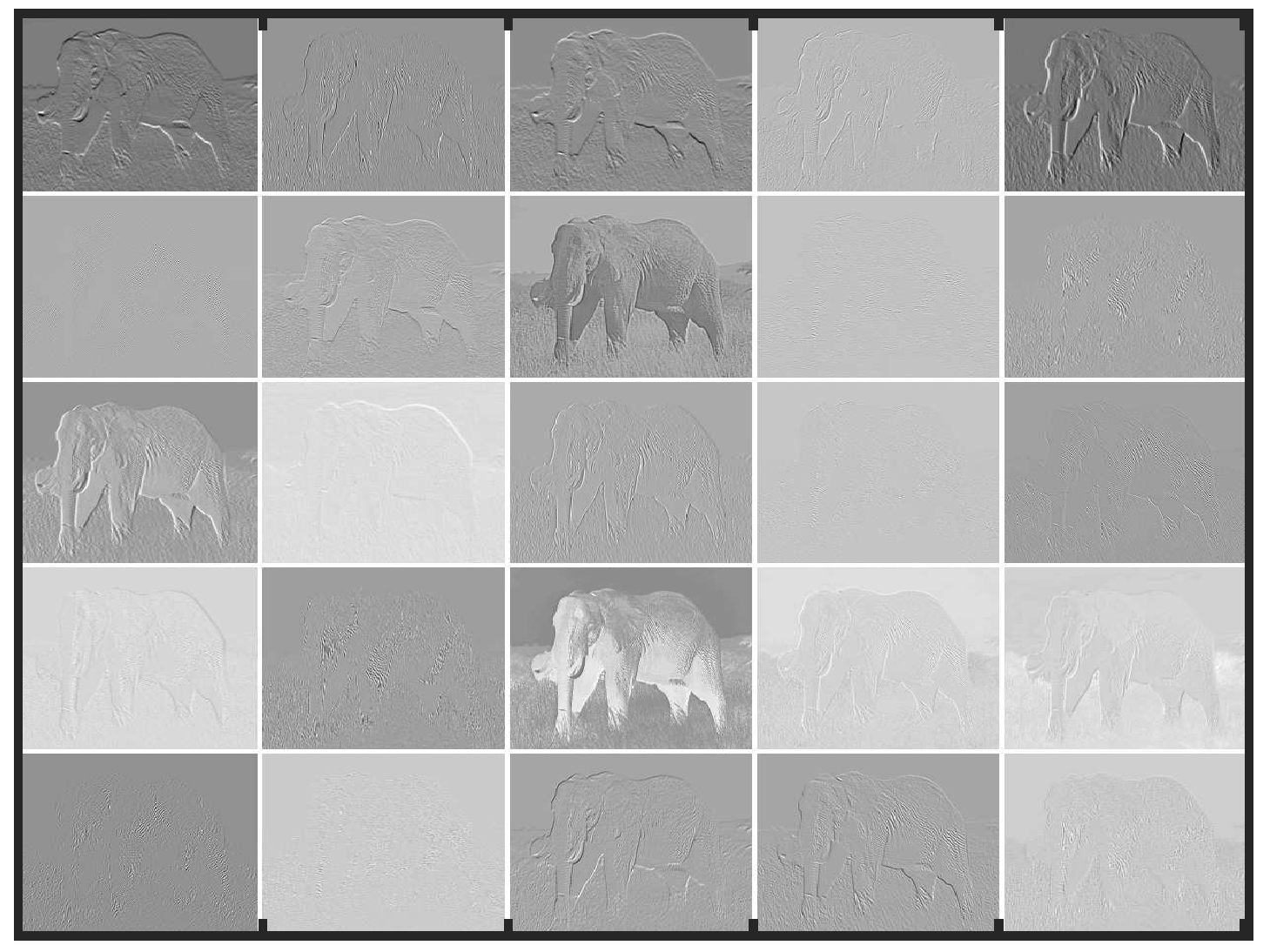}}
\subfigure[Cluster approach]{\label{fig.cluster_cell_image}\includegraphics[width=3in,height=2.5in]{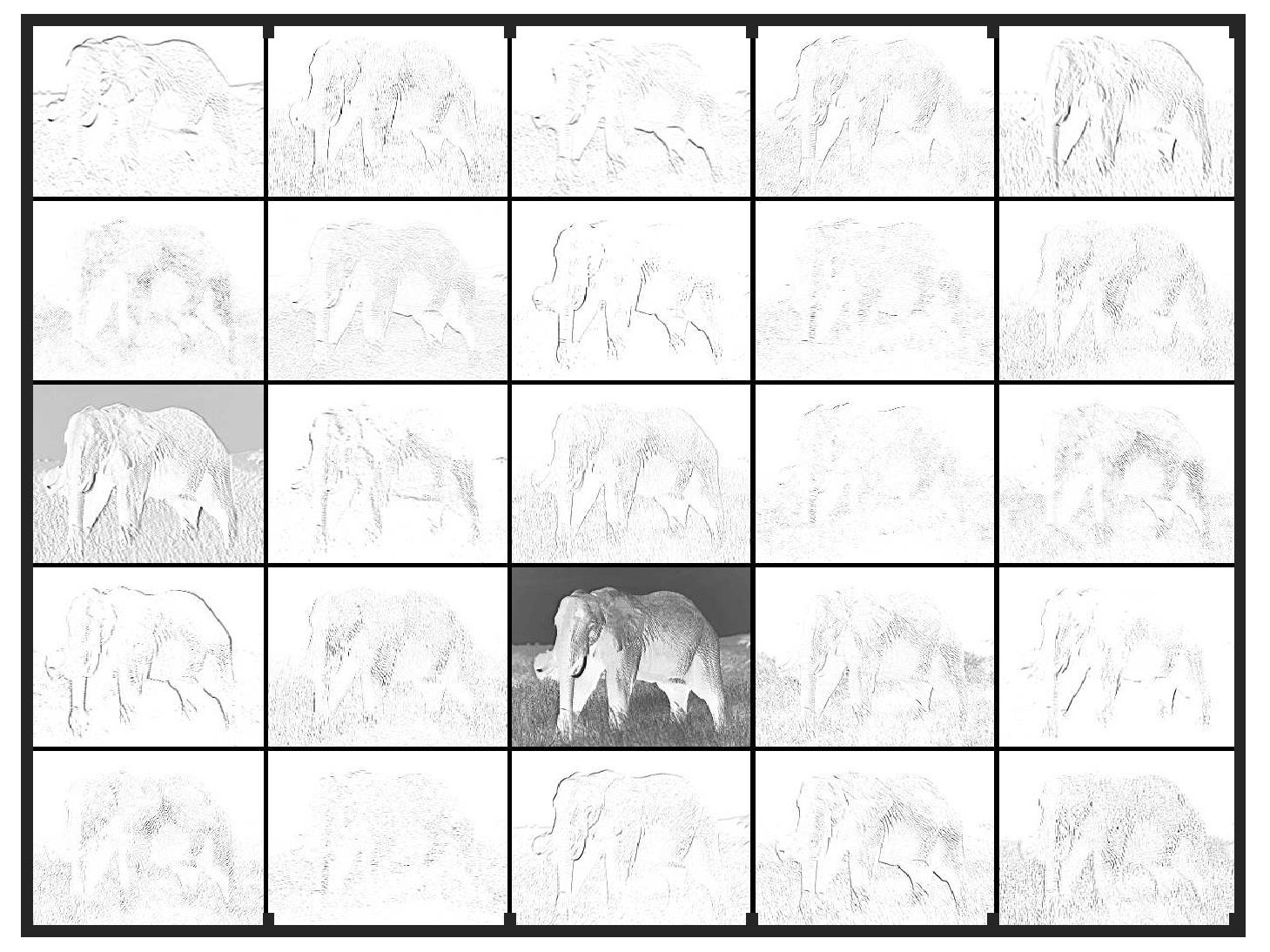}}
\caption{Gray-inverted output of ReLU activation and outputs of RNN convolution operations with the single-cell, twin-cell and cluster approaches in image convolution.}
\end{figure}

\subsection{Numerical Verification of RNN Convolution}
Numerical verification of the RNN convolution approaches is conducted on images (implemented in Theano \cite{2016arXiv160502688short}) by adapting the convolution structure into the RNN.
The convolution kernels used in the experiments are obtained from the first convolution layer of the pre-trained GoogLeNet  \cite{modulegooglenet} provided by \cite{vedaldi15matconvnet}, while the image is obtained from \cite{ILSVRC15}, shown in Figure \ref{fig.original_image}.
The gray-inverted output of a standard convolution operation with the ReLU activation is given in Figure \ref{fig.ReLU_image}.

The outputs in the convolution by using the single-cell, twin-cell and cluster approaches are given in Figures \ref{fig.single_cell_image}, \ref{fig.twin_cell_image} and \ref{fig.cluster_cell_image}. The results demonstrate that these approaches are capable of producing the edge-detection effect similar to the standard convolution.

\subsection{Individual-Cell Based Multi-Layer RNN}

In \cite{yyh_rnnl_ijcnn2017} and Chapter 6 of \cite{phd-yonghua2018}, the authors then present a multi-layer architecture of individual RNN cells described in (\ref{RNN_cell}) (called the MLRNN). 
The MLRNN is then adapted to handle multi-channel datasets (MCMLRNN).
The proposed MLRNN and MCMLRNN achieve similar performance as the MLDRNN and MCMLDRNN presented in Section \ref{chapter.densernn} but cost less computation time.

\subsubsection{Mathematical Model Formulation in Scalar-Form}

\begin{figure}[t]
\centering
\includegraphics[width=3.5in]{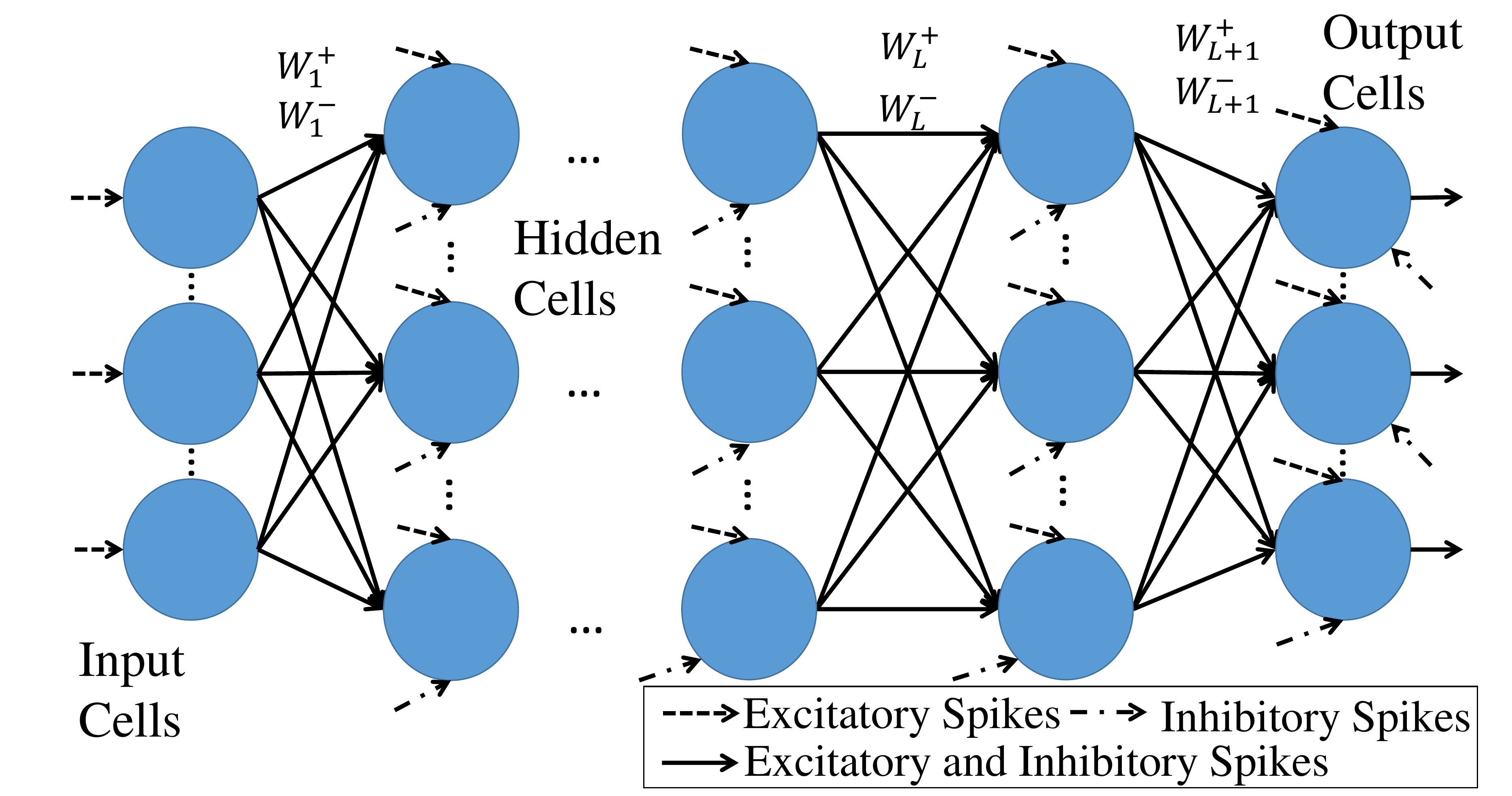}
\caption{Schematic representation of the MLRNN.}
\label{fig.MLIRNN}
\end{figure}

The schematic representation of the multi-layer architecture constituted of individual RNN cells (i.e., the MLRNN) is given in Figure \ref{fig.MLIRNN}.
The MLRNN has $L+2$ layers, where the $l$th ($1\leq l \leq L+2$) layer has $N_l$ cells.\\
1) The first layer is the external-source layer (or say, the input layer), where each cell receives an excitatory spike train from an  external input source. In addition, the cells may fire excitatory and inhibitory spikes towards the cells in the next layer.\\
2) The successive $L$ layers are hidden layers composed of individual RNN cells
that receive both excitatory and inhibitory spike trains from the outside world and cells in the previous layer. Correspondingly, the cells may fire spikes towards the external world and the cells in the next layer.\\
3) The last layer is the output layer, where the RNN cells receive spikes from the outside world and cells in the previous layer and may fire spikes towards the outside world.\\

The following notations are defined for the MLRNN.\\
1) Let $x_{n_1}$ denote the rate of the excitatory spike train from the $n_1$th ($1 \leq n_1 \leq N_1$) external input source.\\
2) Let $q_{n_l}$ ($1 \leq n_l \leq N_l$) denote the stationary excitation probability of the $n_{l}$th  cell in the $l$th layer ($1 \leq l \leq L+2$).\\
3) Let $r_{n_{l}}$ ($1 \leq n_l \leq N_l$) denote the firing rate of the $n_{l}$th cell in the $l$th ($1 \leq l \leq L+2$) layer. For simplicity, set $r_{n_{L+2}}=0$ in the output layer.\\
4) Let $w^+_{n_l, n_{l+1}}=p^+_{n_l, n_{l+1}} r_{n_l}$ and $w^-_{n_l, n_{l+1}}=p^-_{n_l, n_{l+1}} r_{n_l}$ denote excitatory and inhibitory connecting weights between the $n_l$th cell in the $l$th layer and $n_{l+1}$th cell in the $(l+1)$th layer with $l=1,\cdots,L+1$, where $p^+_{n_l, n_{l+1}}$ and $p^-_{n_l, n_{l+1}}$ denote respectively the probabilities of excitatory and inhibitory spikes when the $n_l$th cell in the $l$th layer fires. In addition, let $d_{n_l}$ denote the escape probability of a firing spike. Then, $\sum_{n_{l+1}=1}^{N_{l+1}}(p^+_{n_l, n_{l+1}}+p^-_{n_l, n_{l+1}}) + d_{n_l} = 1$ and $\sum_{n_{l+1}=1}^{N_{l+1}}(w^+_{n_l, n_{l+1}}+w^-_{n_l, n_{l+1}}) + d_{n_l} r_{n_l} =r_{n_l}$. Further, $\sum_{n_{l+1}=1}^{N_{l+1}}(w^+_{n_l, n_{l+1}}+w^-_{n_l, n_{l+1}}) \leq r_{n_l}$.\\
5) Let $\lambda^{+}_{n_{l}}$ and $\lambda^{-}_{n_{l}}$ denote the rates of excitatory and inhibitory spikes from the outside world to the $n_l$th cell in the $l$th-layer  ($1 \leq l \leq L+2$). For the input layer, set $\lambda^{+}_{n_{1}}=\lambda^{-}_{n_{1}}=0$.\\

Based on the spiking RNN theory \cite{RNN89,RNN93}, the stationary excitation probabilities of the MLRNN can be obtained as:
\begin{equation} \label{eqn.MLIRNN_scalar}
\begin{split}
&q_{n_1} = \min(\frac{x_{n_1}}{r_{n_1}},1),\\
&q_{n_l} = \min(\frac{\lambda_{n_l}^{+}+ \sum_{n_{l-1}=1}^{N_{l-1}} q_{n_{l-1}} w^+_{n_{l-1}, n_{l}}}{r_{n_l}+\lambda_{n_l}^{-}+\sum_{n_{l-1}=1}^{N_{l-1}} q_{n_{l-1}} w^-_{n_{l-1}, n_{l}}},1),
\end{split}
\end{equation}
where $l=2,\cdots,L+2.$
The learning parameters (weights and bias) in the MLRNN could be $r_{n_l}$, $\lambda_{n_l}^{+}$, $\lambda_{n_l}^{-}$ $w^+_{n_l, n_{l+1}}$ and $w^-_{n_l, n_{l+1}}$, where the constraints are $r_{n_l}, \lambda_{n_l}^{+}, \lambda_{n_l}^{-}, w^+_{n_l, n_{l+1}}, w^-_{n_l, n_{l+1}} \geq 0$ and $\sum_{n_{l+1}=1}^{N_{l+1}}(w^+_{n_l, n_{l+1}}+w^-_{n_l, n_{l+1}}) \leq r_{n_l}$.

\subsubsection{Mathematical Model Formulation in Matrix-Form}

Suppose there is a dataset represented by a non-negative matrix $X=[x_{d,n_1}] \in \mathbb{R}_{\geq 0}^{D \times N_1}$, where $D$ is the
number of instances, each instance has $N_1$ attributes and $x_{d n_1}$ is the $n_1$th attribute of the $d$th instance.
By import $X$ into the MLRNN, a matrix-form description of the MLRNN (\ref{eqn.MLIRNN_scalar}) can be obtained.\\
1) Let $q_{d, n_l}$ denote the value of $q_{n_l}$ for the $d$th instance. Let a matrix $Q_l=[q_{d, n_l}] \in \mathbb{R}_{\geq 0}^{D \times N_l}$ ($1 \leq l \leq L+2$) denote the value matrix of $q_{d, n_l}$.\\
2) Let a matrix $R_{l}=[r_{d, n_{l}}] \in \mathbb{R}_{\geq 0}^{D \times N_{l}}$ denote the firing-rate matrix for cells in the $l$th layer ($1 \leq l \leq L+2$). Note that, the fire rate of a cell is the same for all instances, i.e., $r_{d_1, n_{l}}=r_{d_2, n_{l}}$ for $1 \leq d_1,d_2 \leq D$.\\
3) Let two matrices $W^{+}_l=[w^+_{n_l, n_{l+1}}] \in \mathbb{R}_{\geq 0}^{N_l \times N_{l+1}}$ and $W^{-}_l=[w^-_{n_l, n_{l+1}}] \in \mathbb{R}_{\geq 0}^{N_l \times N_{l+2}}$ denote excitatory and inhibitory connecting weight matrices between the $l$th and $(l+1)$th layers for $l=1,\cdots,L+1$.\\
4) Let two matrixes $\Lambda_{l}^{+}=[\lambda^{+}_{d,n_{l}}] \in \mathbb{R}_{\geq 0}^{D \times N_{l}}$ and $\Lambda_{l}^{-}=[\lambda^{-}_{d, n_{l}}] \in \mathbb{R}_{\geq 0}^{D \times N_{l}}$ denote the external arrival rate matrices of excitatory and inhibitory spikes for the $l$th layer ($1 \leq l \leq L+2$).
Note also that $\lambda^{+}_{d_1, n_{l}}=\lambda^{+}_{d_2, n_{l}}$ and $\lambda^{-}_{d_1, n_{l}}=\lambda^{-}_{d_2, n_{l}}$  for $1 \leq d_1,d_2 \leq D$.\\
The notations $r_{d, n_{l}}, \lambda^{+}_{d, n_{l}}, \lambda^{-}_{d, n_{l}}$ may be written respectively as $r_{n_{l}}, \lambda^{+}_{n_{l}}, \lambda^{-}_{n_{l}}$ in the rest of the section.
In the matrix form, the excitation probability matrix of the MLRNN can be described as:
\begin{eqnarray} \label{eqn.MLIRNN_matrix}
&&Q_1 =\min(\frac{X}{R_1},1),\\
&&Q_l =\min(\frac{\Lambda_{l}^{+}+Q_{l-1} W^{+}_{l-1}}{R_l+\Lambda_{l}^{-}+Q_{l-1} W^{-}_{l-1}},1),
\end{eqnarray}
where $l=2,\cdots,L+2$ and the division operation between matrices is element-wise.
In addition, $Q_1 \in [0~1]^{D \times N_{1}}$ is the 1st layer output matrix, $Q_l \in [0~1]^{D \times N_{l}}$ is the $l$th layer output ($l=2,\cdots,L+1$) and $Q_{L+2} \in [0~1]^{D \times N_{L+2}}$ (or written as $Q_{L+2}(X)$) is the final MLRNN output matrix.
The parameters in the MLRNN are required to satisfy the RNN constraints: $R_{1},\cdots, R_{L+2} \geq 0$; $\Lambda_{2}^{+},\cdots, \Lambda_{L+2}^{+} \geq 0$; $\Lambda_{2}^{-}, \cdots, \Lambda_{L+2}^{-} \geq 0$; $W^+_{1}, \cdots, W^+_{L+1} \geq 0$; $W^-_{1}, \cdots, W^-_{L+1} \geq 0$ and $\sum_{n_{l+1}=1}^{N_{l+1}}(w^+_{n_l, n_{l+1}}+w^-_{n_l, n_{l+1}}) \leq r_{n_l}$ with $l=1,\cdots, L+1$.

\subsubsection{Learning Problem Description}

Suppose there is a labelled dataset with $D$ instances $\{(X,Y)\}$, where $X =[x_{d, n_{1}}] \in \mathbb{R}_{\geq 0}^{D \times N_{1}}$ is the input-attribute matrix and $Y =[y_{d,n_{L+2}}]\in [0~1]^{D \times N_{L+2}}$ is the desired-output matrix. For the $d$th instance $(x_d,y_d)$, the input-attribute vector is denoted by $x_d = [x_{d,1}~x_{d,2}~\cdots~x_{d,N_{1}}] \in \mathbb{R}^{1 \times N_{1}}$ and the desired-output vector is denoted by $y_d = [y_{d,1}~y_{d,2}~\cdots~y_{d,N_{L+2}}] \in [0~1]^{1 \times N_{L+2}}$. The mapping from $X$ to $Y$ is $f: \mathbb{R}_{\geq 0}^{2 \times N_1} \rightarrow [0~1]^{N_{L+2}}$ and $Y=f(X)$.

The objective is to use the MLRNN to learn the input-output mapping $f$.
In other term, given input $x_d$, the output of the MLRNN $Q_{L+2}(x_d): \mathbb{R}_{\geq 0}^{1 \times N_{1}} \rightarrow [0~1]^{1 \times N_{L+2}}$ should be a reliable and meaningful estimate of the desired output $y_d$. This can be achieved by selecting appropriate firing rates $R_{1},\cdots,R_{L+2} \geq 0$, excitatory-spike arrival rates $\Lambda_{2}^{+}, \cdots, \Lambda_{L+2}^{+} \geq 0$, inhibitory-spike arrival rates $\Lambda_{2}^{-}, \cdots, \Lambda_{L+2}^{-} \geq 0$ and connecting weight matrices $W_1,\cdots,W_{L+1} \geq 0$.

\subsubsection{Training Procedure for MLRNN}

The MLRNN considered here has an input layer, $L$ hidden layers and an output layer ($L+2$ layers in total).
Note that the number of hidden cells (i.e., $N_{L+1}$) in the $(L+1)$th layer needs to be a multiple of 2, which will be explained later in the following subsections.
The training/configuration procedure in this subsection is to find appropriate values for parameters $R_1$ and $R_{l}, \Lambda_{l}^{+}, \Lambda_{l}^{-}, W^+_{l-1}, W^-_{l-1}$ with $l=2,\cdots,L+2$ that satisfy the RNN constraints, so that the MLRNN learns the given dataset $\{(X,Y)\}$.
It is worth pointing out here that the parameters found by the training procedure are required to satisfy the RNN constraints: $R_{1},\cdots, R_{L+2} \geq 0$; $\Lambda_{2}^{+},\cdots, \Lambda_{L+2}^{+} \geq 0$; $\Lambda_{2}^{-}, \cdots, \Lambda_{L+2}^{-} \geq 0$; $W^+_{1}, \cdots, W^+_{L+1} \geq 0$; $W^-_{1}, \cdots, W^-_{L+1} \geq 0$ and $\sum_{n_{l+1}=1}^{N_{l+1}}(w^+_{n_l, n_{l+1}}+w^-_{n_l, n_{l+1}}) \leq r_{d, n_l}$ with $l=1,\cdots, L+1$.
The configuration procedure is presented in the following steps:\\
Step 1. Configure $\Lambda_{l}^{-}$ with $l=2,\cdots,L+2$, $W_{L+1}^{-}$, $R_1$.\\
Step 2. Configure $W^{+}_{l}$, $W^{-}_{l}$, $\Lambda_{l+1}^{+}$, $R_{l+1}$ with $l=1,\cdots,L-1$.\\
Step 3. Configure $W^{+}_{L}$, $W^{-}_{L}$, $\Lambda^{+}_{L+1}$, $R_{L+1}$, $W^{+}_{L+1}$, $\Lambda^{+}_{L+2}$, $R_{L+2}$.

{{\it Step 1. Configure $\Lambda_{l}^{-}$ with $l=2,\cdots,L+2$, $W_{L+1}^{-}$, $R_1$.}
Let us set $\Lambda_{l}^{-}\leftarrow 0$ for $l=2,\cdots,L+2$, $W_{L+1}^{-} \leftarrow 0$ and $R_1 \leftarrow 1$. Since the firing rates of the cells in the input layer are 1 and there is no inhibitory spike from the outside world, then the cell activation (stationary excitation probabaility) in the input layer is quasi-linear (as illustrated in \cite{2016arXiv160908151Y} and Section \ref{chapter.lrnn}), which means $Q_1 =\min(X,1)$ based on (\ref{eqn.MLIRNN_matrix}). For notation ease, let us define the individual-RNN-cell activation $\phi(x^{+},x^{-})|_{\lambda,r}=\min((\lambda+x^{+})/(r+x^{-}),1)$ and use
$\phi(\cdot)$ as a term-by-term function for vectors and matrices.
Then, the system of equations (\ref{eqn.MLIRNN_matrix}) of the MLRNN can be rewritten as
\begin{eqnarray} \label{eqn.MLIRNN_matrix2}
&&Q_1 =\min(X,1),\\
&&Q_l =\phi(Q_{l-1} W^{+}_{l-1},Q_{l-1} W^{-}_{l-1})|_{\Lambda_{l}^{+},R_{l}}~\text{with}~2 \leq l \leq L+1, \\
&&Q_{L+2}=\phi(Q_{L+1} W^{+}_{L+1},0)|_{\Lambda_{L+2}^{+},R_{L+2}}.
\end{eqnarray}
To sum up, in this step, the configurations are conducted via $\Lambda_{l}^{-}\leftarrow 0$ for $l=2,\cdots,L+2$, $W_{L+1}^{-} \leftarrow 0$ and $R_1 \leftarrow 1$.

{\it Step 2. Configure $W^{+}_{l}$, $W^{-}_{l}$, $\Lambda_{l+1}^{+}$, $R_{l+1}$ with $l=1,\cdots,L-1$.} \label{secmlrnn.conf_W1_L}
First, let us set $W^{+}_{l} \leftarrow 0$ with $l=1,\cdots,L-1$. Then, a series of reconstruction problems related to the cell activations of the current layer and $W_{l}^{-}$ are constructed and solved for $W_{l}^{-}$ with $l=1,\cdots,L-1$.
Specifically, the reconstruction problem for the weight matrix $W_1^{-}$ is constructed as
\begin{equation} \label{eqn.mlrnn_w1}
\begin{split}
& \min_{W_1}  ||X_1 -  \sigma(\phi(0,X_1 \bar{W})|_{\max(X_1 \bar{W}),\max(X_1 \bar{W})}) W_{1}^- ||^2 +||W_{1}^-||_{\ell_1},~  \text{s.t.~}W_{1}^- \geq 0,
\end{split}
\end{equation}
where $\bar{W} \geq 0$ is a randomly-generated matrix with appropriate dimensions that satisfies the RNN constraints and operation $\max(\cdot)$ produces the maximal element of its input.
Besides, the operation $\sigma(H)$ first maps each column of its input $H$ into $[0~1]$ linearly, then uses the ``zcore'' MATLAB operation and finally adds a positive constant to remove negativity. Note that $X_1$ is obtained via $X_1 \leftarrow X$. In addition, the fast iterative shrinkage-thresholding algorithm (FISTA) in Chapter \ref{chapter.densernn}, which has been adapted from \cite{FISTA} with the modification of setting negative elements in the solution to zero in each iteration, is exploited to solve the reconstruction problem (\ref{eqn.mlrnn_w1}).
After solving the problem, the values of $W_{1}^{-} \geq 0$ are obtained and then adjusted to satisfy the RNN constraints $\sum_{n_{2}=1}^{N_{2}}(w^+_{n_1, n_{2}}+w^-_{n_1, n_{2}}) = \sum_{n_{2}=1}^{N_{2}} w^-_{n_1, n_{2}} \leq 1$. If $\sum_{n_{2}=1}^{N_{2}} w^-_{n_1, n_{2}} > 1$, the operation $w^-_{n_1, n_{2}} \leftarrow w^-_{n_1, n_{2}}/ \left(\sum_{n_{2}=1}^{N_{2}} w^-_{n_1, n_{2}}\right)$ with $n_2=1,\cdots,N_{2}$ can be used to guarantee that the weights satisfy the RNN constraints. Then, the external arrival rates and firing rates of the cells in the next layer are set via $\Lambda_{2}^{+} \leftarrow \max(X_1 W^{-}_{1})/5$ and $R_{2} \leftarrow \max(X_1 W^{-}_{1})/5$.

In sequence, the following reconstruction problem is solved for $W_{l}^{-}$ with $l=2,\cdots,L-1$ using the modified FISTA:
\begin{equation} \label{eqn.mlrnn_wl}
\begin{split}
& \min_{W^{+}_{l}} ||X_l -  \sigma(\phi(0, X_l \bar{W})|_{\max(X_l \bar{W}),\max(X_l \bar{W})}) W_{l}^{-} ||^2+||W_{l}^{-}||_{\ell_1},~ \text{s.t.~}W_{l}^{-} \geq 0,
\end{split}
\end{equation}
where $X_l$ is its layer encodings obtained via $X_l \leftarrow \phi(0, X_{l-1} W_{l-1})|_{\Lambda_{l}^{+},R_{l}}$ and $\bar{W} \geq 0$ is randomly generated and adjusted to satisfy the RNN constraints.
After solving the problem, the matrix $W_{l}^{-} \geq 0$ is obtained; and then its values are adjusted to satisfy the RNN constraints $\sum_{n_{l+1}=1}^{N_{l+1}}(w^+_{n_l, n_{l+1}}+w^-_{n_l, n_{l+1}}) = \sum_{n_{l+1}=1}^{N_{l+1}} w^-_{n_l, n_{l+1}} \leq r_{ n_{l}}$ by using the operation $w^-_{n_l, n_{l+1}} \leftarrow w^-_{n_l, n_{l+1}}/ \left( (\sum_{n_{l+1}=1}^{N_{l+1}} w^-_{n_l, n_{l+1}})/r_{ n_{l}}\right)$ with $n_{l+1}=1,\cdots,N_{l+1}$. Then, the external arrival rates and firing rates of the cells in the next layer are set via $\Lambda_{l+1}^{+} \leftarrow \max(X_l W^{-}_{l})/5$ and $R_{l+1} \leftarrow \max(X_l W^{-}_{l})/5$ with $l=2,\cdots,L-1$.

To sum up, in this step, the configurations are conducted as follows: 1) $W^{+}_{l} \leftarrow 0$ with $l=1,\cdots,L-1$; 2) $W^{-}_{l}$ with $l=1,\cdots,L-1$ are configured by solving reconstruction problems (\ref{eqn.mlrnn_w1}) and (\ref{eqn.mlrnn_wl}); 3) $\Lambda_{l+1}^{+}$, $R_{l+1}$ are configured via $\Lambda_{l+1}^{+} \leftarrow \max(X_l W^{-}_{l})/5$ and $R_{l+1} \leftarrow \max(X_l W^{-}_{l})/5$ with $l=1,\cdots,L-1$.

{\it Step 3. Configure $W^{+}_{L}$, $W^{-}_{L}$, $\Lambda^{+}_{L+1}$, $R_{L+1}$, $W^{+}_{L+1}$, $\Lambda^{+}_{L+2}$, $R_{L+2}$.}
A single-hidden-layer RNN-based artificial neural network (SLANN) is first constructed using the ELM concept \cite{elm,mlelm} (with $N_{L+1}/2$ hidden units and $N_{L+2}$ output units);  and then its weights are  utilized to configure
$W^{+}_{L}$, $W^{-}_{L}$, $\Lambda^{+}_{L+1}$, $R_{L+1}$, $W^{+}_{L+1}$, $\Lambda^{+}_{L+2}$, $R_{L+2}$.

For this SLANN whose activation function is $\bar{\phi}(x)|_{\alpha}=\alpha/(\alpha+x)$ with parameter $\alpha>0$ to be determined, the input and output weights are denoted by $\bar{W}_1=[\bar{w}_{n_{L},\bar{n}_1}] \in \mathbb{R}_{\geq 0}^{N_{L} \times (N_{L+1}/2)}$ and $\bar{W}_2 =[\bar{w}_{\bar{n}_1,n_{L+2}}] \in \mathbb{R}^{(N_{L+1}/2) \times N_{L+2}}$.
Let $X_{L}$ denote the $L$th-layer output of the MLRNN and $Y$ is the desired output (or say, labels corresponding to the training dataset). Based on $W_{L-1}^{-}$ configured in Step 2,
we have $X_{L} \leftarrow \phi(0, X_{L-1} W_{L-1})|_{\Lambda^{+}_{L},R_{L}}$.
With $X_L$ imported into the SLANN, a forward pass can be described as $\bar{O}(X_L) = \bar{\phi}(X_L \bar{W}_1)|_{\alpha} \bar{W}_2: \mathbb{R}^{D \times N_L} \rightarrow \mathbb{R}^{D \times N_{L+2}}$.\\
The input weights $\bar{W}_1$ are randomly generated in range $[0 ~ 1]$ and then normalized
to satisfy the RNN constraints $2 \sum_{\bar{n}_{1}=1}^{N_{L+1}/2} \bar{w}_{n_{L}, \bar{n}_1} \leq r_{n_{L}}$ by using the operation $\bar{w}_{n_{L},\bar{n}_1} \leftarrow \bar{w}_{n_{L},\bar{n}_1}/ \left(2\sum_{\bar{n}_{1}=1}^{N_{L+1}/2} \bar{w}_{n_{L}, \bar{n}_1} \right)$ with $n_{L}=1,\cdots,N_{L}$.
The parameter $\alpha$ in the activation function is set by $\alpha \leftarrow \max(X_L \bar{W}_1)/5$.
Then, the output weights $\bar{W}_2$ are determined using the Moore-Penrose pseudo-inverse \cite{gelenbedeep2016,zhang2014cross,yin2012weights,zhang2012pruning,elm,mlelm} (denoted by ``pinv'') as:
\begin{equation}
\bar{W}_2 \leftarrow \text{pinv}(\bar{\phi}(X_L \bar{W}_1)|_{\alpha}) Y.
\end{equation}
Then, we adjust $\bar{W}_2$ via $\bar{W}_2 \leftarrow \bar{W}_2/\text{sum}(|\bar{W}_2|)$ to make the summation of all elements in $|\bar{W}_2|$ no larger than 1, where $\text{sum}(\cdot)$ produces the summations of all elements of its input.

The following shows how to configure $W^{+}_{L}$, $W^{-}_{L}$, $\Lambda^{+}_{L+1}$, $R_{L+1}$, $W^{+}_{L+1}$, $\Lambda^{+}_{L+2}$, $R_{L+2}$ in the MLRNN by using $\bar{W}_1$, $\alpha=\max(X_L \bar{W}_1)/5$ and $\bar{W}_2$ in the SLANN.
For illustration ease, let us define $\hat{\phi}(x)|_{\alpha}=x/(\alpha+x)$. It is evident that $\bar{\phi}(x)|_{\alpha}=1-\hat{\phi}(x)|_{\alpha}$.
Let $\bar{W}^{+}_2 =\max(\bar{W}_2 ,0) \geq 0$ and $\bar{W}^{-}_2 =\max(-\bar{W}_2 ,0) \geq 0$. Since the output matrix of the SLANN is $\bar{O}(X_L) = [\bar{o}_{d,n_{L+2}}]=\bar{\phi}(X_L \bar{W}_1)|_{\alpha} \bar{W}_2  \in \mathbb{R}^{D \times N_{L+2}}$, then the $n_{L+2}$th ($n_{L+2}=1,\cdots,N_{L+2}$) output of this SLANN for the $d$th instance is
\begin{equation*}
\begin{split}
&\bar{o}_{d,n_{L+2}}= \sum_{\bar{n}_1=1}^{N_{L+1}/2}\left((\bar{\phi}(X \bar{W}_1)|_{\alpha})_{d,\bar{n}_1} (\bar{W}_2)_{\bar{n}_1, n_{L+2}}\right) \\
&= \sum_{\bar{n}_1=1}^{N_{L+1}/2}(\bar{\phi}(X \bar{W}_1)|_{\alpha})_{d,\bar{n}_1} (\bar{W}^{+}_2)_{\bar{n}_1, n_{L+2}} -\sum_{\bar{n}_1=1}^{N_{L+1}/2}(\bar{\phi}(X \bar{W}_1)|_{\alpha})_{d, \bar{n}_1} (\bar{W}^{-}_2)_{\bar{n}_1, n_{L+2}}\\
&= \sum_{\bar{n}_1=1}^{N_{L+1}/2}(\bar{\phi}(X \bar{W}_1)|_{\alpha})_{d,\bar{n}_1} (\bar{W}^{+}_2)_{\bar{n}_1, n_{L+2}} +\sum_{\bar{n}_1=1}^{N_{L+1}/2}(\hat{\phi}(X \bar{W}_1)|_{\alpha})_{d,\bar{n}_1}(\bar{W}^{-}_2)_{\bar{n}_1, n_{L+2}}-\sum_{\bar{n}_1=1}^{N_{L+1}/2} (\bar{W}^{-}_2)_{\bar{n}_1, n_{L+2}}.
\end{split}
\end{equation*}
Since $\bar{\phi}(X \bar{W}_1)|_{\alpha}=\phi(0,X \bar{W}_1)|_{\alpha,\alpha}$ and $\hat{\phi}(X \bar{W}_1)|_{\alpha}=\phi(X \bar{W}_1,X \bar{W}_1)|_{0,\alpha}$, then
\begin{equation*}
\begin{split}
&\bar{o}_{d,n_{L+2}}= \sum_{\bar{n}_1=1}^{N_{L+1}/2}(\phi(0,X \bar{W}_1)|_{\alpha,\alpha})_{d,\bar{n}_1} (\bar{W}^{+}_2)_{\bar{n}_1, n_{L+2}} \\
& +\sum_{\bar{n}_1=1}^{N_{L+1}/2}(\phi(X \bar{W}_1,X \bar{W}_1)|_{0,\alpha})_{d,\bar{n}_1}(\bar{W}^{-}_2)_{\bar{n}_1, n_{L+2}}-\sum_{\bar{n}_1=1}^{N_{L+1}/2} (\bar{W}^{-}_2)_{\bar{n}_1, n_{L+2}}.
\end{split}
\end{equation*}

For the MLRNN, let us configure $\Lambda_{L+2}^{+}=[\lambda^{+}_{d,n_{L+2}}] \in \mathbb{R}_{\geq 0}^{D \times N_{L+2}}$ in the $(L+2)$th layer via $\lambda^{+}_{d, n_{L+2}} \leftarrow \max(\text{sum}(\bar{W}^{-}_2,1)) - \sum_{\bar{n}_1=1}^{N_{L+1}/2} (\bar{W}^{-}_2)_{\bar{n}_1, n_{L+2}}$ for $d=1,\cdots,D$, where operation $\text{sum}(\bar{W}^{-}_2,1)$ produces the summation vector of each column in matrix $\bar{W}^{-}_2$.

The configurations of $\Lambda_{L+1}^{+}=[\lambda^{+}_{d,n_{L+1}}] \in \mathbb{R}_{\geq 0}^{D \times N_{L+1}}$ and $R_{L+1}=[r_{d,n_{L+1}}] \in \mathbb{R}_{\geq 0}^{D \times N_{L+1}}$ are divided into two parts.
From the $1$st to $(N_{L+1}/2)$th hidden cells in the $(L+1)$th layer, $\lambda^{+}_{d,n_{L+1}} \leftarrow \alpha$ and $r_{d,n_{L+1}} \leftarrow \alpha$ with $n_{L+1}=1,\cdots,N_{L+1}/2$; while, for the $(N_{L+1}/2+1)$th to $N_{L+1}$th hidden cells, $\lambda^{+}_{d,n_{L+1}} \leftarrow 0$ and $r_{d,n_{L+1}} \leftarrow \alpha$ with $n_{L+1}=N_{L+1}/2+1,\cdots,N_{L+1}$, where $d=1,\cdots,D$.

The connecting weights $W_{L}^{+}=[w_{n_{L},n_{L+1}}^{+}] \in \mathbb{R}_{\geq 0}^{N_{L},N_{L+1}}$, $W_{L}^{-}=[w_{n_{L},n_{L+1}}^{-}]\in \mathbb{R}_{\geq 0}^{N_{L},N_{L+1}}$ and $W_{L+1}^{+}=[w_{n_{L+1},n_{L+2}}^{+}] \in \mathbb{R}_{\geq 0}^{N_{L+1},N_{L+2}}$ are configured  based on $\bar{W}_1$, $\bar{W}_2^{+}$ and $\bar{W}_2^{-}$, which are also divided into two parts.\\
First, let us set $w_{n_{L},\bar{n}_1}^{+} \leftarrow 0$, $w_{n_{L},\bar{n}_1}^{-} \leftarrow \bar{w}_{n_L,\bar{n}_1}$ and $w_{\bar{n}_1, n_{L+2}}^{+} \leftarrow \bar{w}_{\bar{n}_1, n_{L+2}}^{+}$ with $n_{L}=1,\cdots,N_{L}$, $\bar{n}_{1}=1,\cdots,N_{L+1}/2$ and $n_{L+2}=1,\cdots,N_{L+2}$.\\
Then, let us set $w_{n_{L},\bar{n}_1}^{+} \leftarrow \bar{w}_{n_L,\bar{n}_1}$, $w_{n_{L},\bar{n}_1}^{-} \leftarrow \bar{w}_{n_L,\bar{n}_1}$ and $w_{\bar{n}_1, n_{L+2}}^{+} \leftarrow \bar{w}_{\bar{n}_1, n_{L+2}}^{-}$ with $n_{L}=1,\cdots,N_{L}$, $\bar{n}_{1}=N_{L+1}/2+1,\cdots,N_{L+1}$ and $n_{L+2}=1,\cdots,N_{L+2}$.\\
Simply put,
$W^{+}_{L}\leftarrow [\textbf{0}; \bar{W}_1]$, $W^{-}_{L}\leftarrow [\bar{W}_1; \bar{W}_1]$ and $W^{+}_{L+1} \leftarrow [\bar{W}^{+}_2;\bar{W}^{-}_2]$, where $\textbf{0}$ is an all-zero matrix with appropriate dimensions.

After the configurations, the final output $Q_{L+2} = [q_{d,n_{L+2}}] \in [0~1]^{D \times N_{L+2}}$ of the MLRNN can be obtained as
\begin{equation}
q_{d,n_{L+2}} = \bar{o}_{d,n_{L+2}} + \max(\text{sum}(\bar{W}^{-}_2,1)),
\end{equation}
with $d=1,\cdots,D$ and $n_{L+2}=1,\cdots,N_{L+2}$.

\begin{figure}[t]
\centering
\includegraphics[width=3.5in]{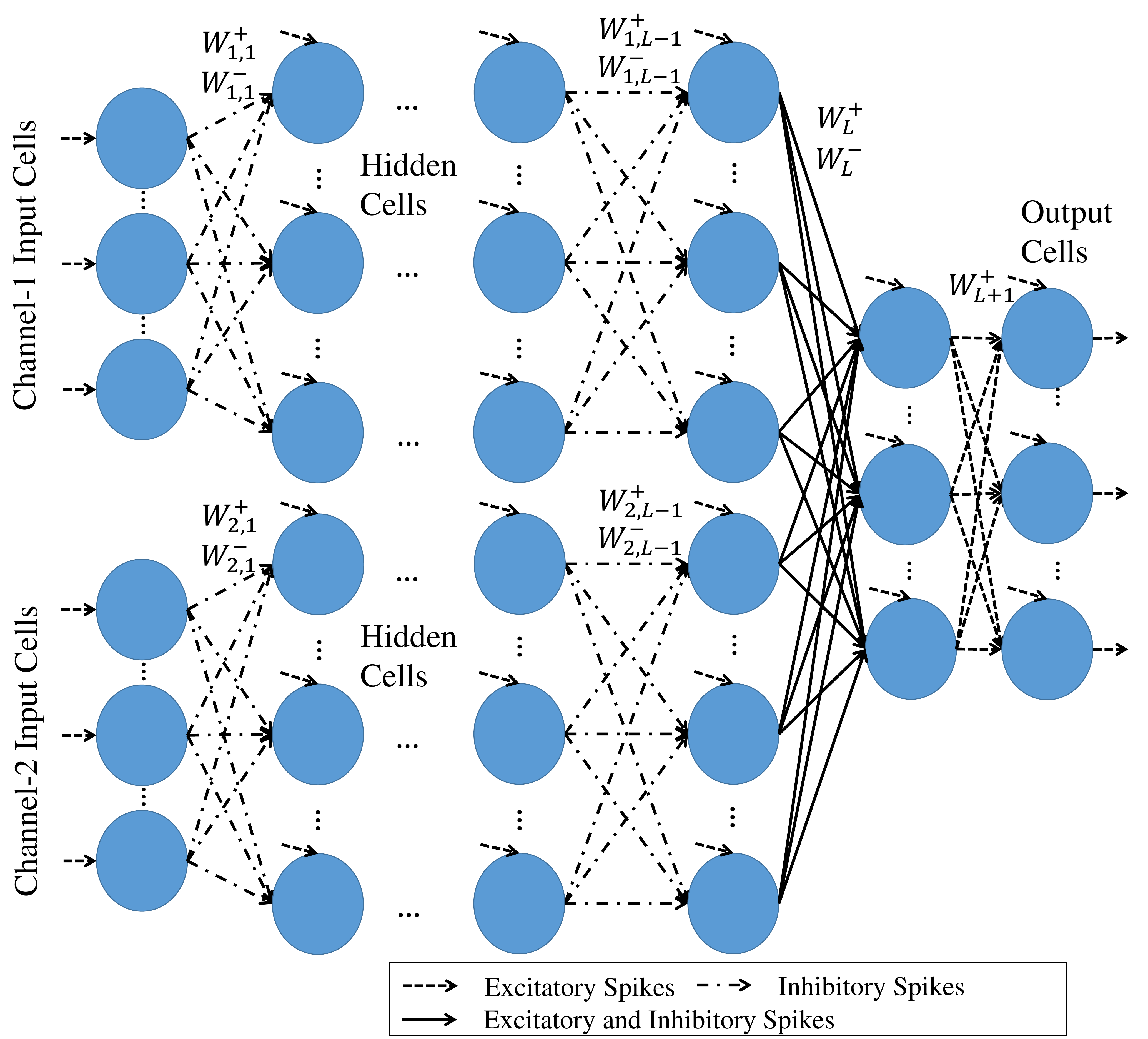}
\caption{Schematic representation of the MCMLRNN.}
\label{fig.MCFRNN}
\end{figure}

\subsubsection{Multi-Channel MLRNN}

The MLRNN can be adapted to handle multi-channel datasets (denoted as the MCMLRNN), shown in Figure \ref{fig.MCFRNN}, where the connecting weights between layers for only Channel-$c$ ($c=1,\cdots,C$) are $W^{+}_{c,l},W^{-}_{c,l}\geq 0$ ($l=1,\cdots,L-1)$, those between the $(L-1)$th and $L$th hidden layers are $W^{+}_{L},W^{-}_{L}\geq 0$ and output weights are $W^{+}_{L+1}\geq 0$. Besides, a vector $\Lambda^{+}$ denotes the external arrival rates of excitatory spikes for the cells in the $(L+2)$th layer, which is the output layer. The training procedure of the MCMLRNN can be generalized from that of the MLRNN.

\begin{table*}[t]
\caption{Testing accuracies ($\%$), training and testing time (s) of different methods for DAS dataset.} \label{tab.sport}
\begin{center}
\begin{tabular}{|l|l|l|l|}
\hline
Method  &Testing accuracy &Training time &Testing time 	\\ \hline
MCMLRNN&\textbf{99.21} &12.38&0.96\\	\hline
MCMLDRNN\cite{yin2016deep}&\textbf{99.21} &26.81&4.97\\	\hline
MCMLDRNN1\cite{yin2016deep}&98.98&89.16&12.86\\	\hline
MCMLDRNN2\cite{yin2016deep}&94.67&177.03&10.88\\	\hline
Improved MLDRNN\cite{yin2016deep}&92.17&13.11&1.09\\	\hline
Original MLDRNN \cite{gelenbedeep2016}&92.83&6.02&0.74\\	\hline
MLP$+$dropout \cite{chollet2015keras} &91.94&3291.47&0.50	\\	\hline
CNN \cite{chollet2015keras} &98.52&1289.76&0.47\\	\hline
CNN$+$dropout \cite{chollet2015keras} &99.05&1338.35&0.53\\	\hline
H-ELM  \cite{mlelm} &96.58&9.60&0.82\\	\hline
\end{tabular}
\end{center}
\end{table*}

\subsection{Numerical Results}

The numerical tests for the MLRNN and MCMLRNN are conducted using three multi-channel classification datasets: an image dataset and two real-world time-series datasets, which are the small NORB dataset \cite{lecun2004learning}, DSA dataset \cite{altun2010comparative,barshan2014recognizing,altun2010human} and Twin Gas Sensor Arrays (TGSA) dataset \cite{fonollosa2016calibration}.
The proposed MCMLRNN is compared with the  MLDRNN presented in \cite{yin2016deep,gelenbedeep2016} and Section \ref{chapter.densernn}, the multi-layer perception (i.e., the MLP) \cite{chollet2015keras}, the convolutional neural network (i.e., the CNN) \cite{chollet2015keras,srivastava2014dropout} and hierarchical
ELM (i.e., the H-ELM) \cite{mlelm}.
The corresponding numerical results well demonstrate that the MCMLRNN is effective and the most efficient among the compared deep-learning tools, as well as the value of individual RNN cells for deep learning.
For example, Table \ref{tab.sport} presents the testing accuracies, training and testing time of different methods for classifying the DAS dataset. It can be seen from the table that both the MCMLRNN and MCMLDRNN achieve the highest testing accuracies among all deep-learning tools and have higher training efficiency than the CNN equipped with the dropout technique that achieves the third highest testing accuracy. Comparing with the MCMLDRNN, the MCMLRNN has a lower computational complexity in terms of both training and testing. More related results can be found in  \cite{yyh_rnnl_ijcnn2017} and Chapter 6 of \cite{phd-yonghua2018}.

\section{Conclusions and Future Work}
\label{ch:conclusions}

\subsection{Summary of Achievements}

This paper has presented recent progresses on connecting the RNN methods and deep learning in \cite{gelenbedeep2016,gelenbedeep2016_SAI,yin2016deep,
yin_cloud2017,2016arXiv160908151Y,gelenbe2017deep,yyh_rnnl_ijcnn2017,phd-yonghua2018} since the first related work was published in 2016 \cite{gelenbedeep2016}:\\
1) Based on the RNN function approximator proposed in 1999 \cite{gelenbe1999function,99RNN_appro,gelenbe2004function}, the approximation capability of the RNN has been investigated and an efficient classifier has been developed.\\
2) Multi-layer non-negative RNN autoencoders have been proposed for robust dimension reduction.\\
3) The model of the dense RNN that incorporates both soma-to-soma interactions and conventional synaptic connections among neurons has been proposed and an efficient deep-learning tool based on the dense RNN has been developed.\\
4) The power of the standard RNN for deep learning has been investigated. The ability of the RNN with only positive parameters to conduct image convolution operations has been demonstrated. The MLRNN equipped with the developed training algorithm achieves comparable or better classification at a lower computation cost than conventional DL methods.

The achievements in the RNN methods and deep learning can be summarised into the following aspects:\\
1) The properties of the random neural networks have been investigated from different aspects, such as its approximation properties, various types of structures, and numerous learning strategies.\\
2) Based on these RNN-property investigations, the RNN has been successfully connected with deep learning and the capabilities of the RNN as a deep-learning tool are well investigated.\\
3) Efficient and effective neural-network learning tools based on the RNN have been developed for handling various data challenges arising from various systems, such as the over-fitting effect, diversified attribute characteristics, speed requirements, and so forth.\\

\subsection{Future Work}

In the directions of deep learning with the RNN, future work could focus on the following aspects:
\begin{enumerate}
\item
The capabilities of the deep-learning tools based on the RNN and dense RNN could be further evaluated in more practical applications when solving real-world problems.
\item
Extensive comparisons of the proposed RNN tools with other related machine-learning tools could be further investigated.
\item
The training and testing phases of the multi-layer RNN could be integrated into the existing deep-learning platforms, such as Tensorflow \cite{tensorflow2015-whitepaper}.
\item
The feasibility of applying the RNN theory to modelling the existing neuromorphic computing hardware platforms may be worthy of investigation.
\item
The approach to utilising linearisation facilitates the analysis of deep neural networks from a theoretical perspective \cite{phd-yonghua2018}, and as such this may be worth investigating further in order to better understand the deep neural networks from the theoretical perspective and develop improved deep-learning tools.
\item
The size of dataset in the issues of deep learning can be very large. Transforming the deep-learning problem into a moment-learning problem \cite{phd-yonghua2018} using the statistical estimations of raw moments from the dataset could significantly reduce the problem's complexity. Thus, this could be a promising direction for future work on deep learning.
\end{enumerate}

\section*{Acknowledgement}
The author would like to thank the support of the President's PhD
Scholarship from Imperial College London.

\bibliographystyle{IEEEtran}
\bibliography{RNNDL}

\end{document}